%% file: main.tex
\documentclass{article}

\usepackage{iclr2027_conference,times}
\usepackage[utf8]{inputenc}
\usepackage[T1]{fontenc}
\usepackage{amsmath,amssymb,amsfonts}
\usepackage{algorithm}
\usepackage{algpseudocode}
\usepackage{booktabs}
\usepackage{graphicx}
\usepackage{longtable}
\usepackage{microtype}
\usepackage{tikz}
\usepackage{xcolor}
\usepackage{url}
\usepackage[colorlinks=true,allcolors=blue!55!black]{hyperref}
\usetikzlibrary{arrows.meta,positioning}

\title{Calibration-Risk Routing for Controlled\\
World-Model Adaptation}

\author{Yifan Zhang$^{1}$ \quad Liang Zheng$^{1}$ \quad Qifan Zhang$^{2}$\\
\normalfont $^{1}$Central South University\\
$^{2}$Faculty of Information Technology, Monash Malaysia\\
{\small\texttt{erzhu419@gmail.com}, \texttt{204201048@csu.edu.cn}}\\
{\small\texttt{zhengliang@csu.edu.cn},
\href{mailto:zhangqifan0419@gmail.com}{\texttt{zhangqifan0419@gmail.com}}}}

\iclrfinalcopy

\begin{document}
\maketitle
\lhead{}
\begin{center}
\small Code: \url{https://github.com/erzhu419/MC-WM}
\end{center}
\input{generated/final_confirmation/algorithm_confirmation_results_macros.tex}
\input{generated/final_exact_routing/exact_routing_results_macros.tex}

\begin{abstract}
Model-based reinforcement learning (MBRL) can exploit simulated experience,
but a simulator-to-target shift creates a model-selection problem: correcting
the simulator and fitting the target directly can each fail under limited target
data. We introduce the Model-Corrected World Model (MC-WM), which separates
initial target data into disjoint fit, selection, and calibration partitions and deploys
the family with lower standardized calibration risk. A learned confidence
signal and deterministic validity predicates weight one-step imagined policy
updates without rewriting physical rewards. We evaluate 540 unique reported run
cells across three controlled Multi-Joint dynamics with Contact (MuJoCo)
shifts; one exact-routing cell was repeated after a pre-deployment artifact
gate, giving 541 completed executions. In a fresh
ten-seed confirmation, MC-WM exceeds target Soft Actor-Critic (SAC) return by
$\FinalAntCandidateOverTargetSACReturnDelta$
[$\FinalAntCandidateOverTargetSACReturnCILow$,
$\FinalAntCandidateOverTargetSACReturnCIHigh$],
$\FinalGravityCandidateOverTargetSACReturnDelta$
[$\FinalGravityCandidateOverTargetSACReturnCILow$,
$\FinalGravityCandidateOverTargetSACReturnCIHigh$], and
$\FinalWalkerCandidateOverTargetSACReturnDelta$
[$\FinalWalkerCandidateOverTargetSACReturnCILow$,
$\FinalWalkerCandidateOverTargetSACReturnCIHigh$] on Ant actuator, HalfCheetah
gravity, and Walker friction, respectively, while reducing violations in all
three. It also exceeds matched Domain Adaptation with Rewards from Classifiers
(DARC) return in every environment, but its
return differences from direct target Model-Based Policy Optimization (MBPO)
remain unresolved. The selector chooses direct target models for all Ant and
gravity runs and residual models for all Walker runs. Automatic routing exactly
reproduces its selected forced arm in all 30 paired checks and improves Ant
return over the unselected arm by $\ExactRoutingAntReturnDelta$
[$\ExactRoutingAntReturnCILow$, $\ExactRoutingAntReturnCIHigh$]; gravity and
Walker return and all route-level violation differences remain unresolved. A separate
sparse-versus-dense study finds fewer residual parameters but no resolved
endpoint advantage. The evidence supports calibrated model-family choice and
bounded trust mechanisms, not universal residual or symbolic superiority.
\end{abstract}

\input{sections/introduction}
\input{sections/related_work}
\input{sections/method}
\input{sections/experiments}
\input{sections/results}
\input{sections/discussion}
\input{sections/statements}

\bibliography{references}
\bibliographystyle{iclr2027_conference}

\appendix
\input{sections/appendix}

\end{document}

%% file: generated/final_confirmation/algorithm_confirmation_results_macros.tex
\newcommand{\FinalAntDARCReturnMean}{1862.2}
\newcommand{\FinalAntDARCReturnCILow}{1566.3}
\newcommand{\FinalAntDARCReturnCIHigh}{2197.4}
\newcommand{\FinalAntDARCViolationMean}{1.80}
\newcommand{\FinalAntDARCViolationCILow}{0.59}
\newcommand{\FinalAntDARCViolationCIHigh}{3.35}
\newcommand{\FinalAntDirectMBPOReturnMean}{4011.5}
\newcommand{\FinalAntDirectMBPOReturnCILow}{2229.7}
\newcommand{\FinalAntDirectMBPOReturnCIHigh}{6129.5}
\newcommand{\FinalAntDirectMBPOViolationMean}{1.26}
\newcommand{\FinalAntDirectMBPOViolationCILow}{0.59}
\newcommand{\FinalAntDirectMBPOViolationCIHigh}{2.13}
\newcommand{\FinalAntCandidateReturnMean}{4084.9}
\newcommand{\FinalAntCandidateReturnCILow}{2917.1}
\newcommand{\FinalAntCandidateReturnCIHigh}{5377.8}
\newcommand{\FinalAntCandidateViolationMean}{1.39}
\newcommand{\FinalAntCandidateViolationCILow}{0.73}
\newcommand{\FinalAntCandidateViolationCIHigh}{2.10}
\newcommand{\FinalAntCoreReturnMean}{1361.7}
\newcommand{\FinalAntCoreReturnCILow}{1061.1}
\newcommand{\FinalAntCoreReturnCIHigh}{1671.0}
\newcommand{\FinalAntCoreViolationMean}{1.38}
\newcommand{\FinalAntCoreViolationCILow}{0.57}
\newcommand{\FinalAntCoreViolationCIHigh}{2.31}
\newcommand{\FinalAntRawSimReturnMean}{1588.6}
\newcommand{\FinalAntRawSimReturnCILow}{1376.7}
\newcommand{\FinalAntRawSimReturnCIHigh}{1860.7}
\newcommand{\FinalAntRawSimViolationMean}{0.41}
\newcommand{\FinalAntRawSimViolationCILow}{0.20}
\newcommand{\FinalAntRawSimViolationCIHigh}{0.62}
\newcommand{\FinalAntTargetSACReturnMean}{2159.2}
\newcommand{\FinalAntTargetSACReturnCILow}{1895.6}
\newcommand{\FinalAntTargetSACReturnCIHigh}{2436.1}
\newcommand{\FinalAntTargetSACViolationMean}{3.95}
\newcommand{\FinalAntTargetSACViolationCILow}{3.03}
\newcommand{\FinalAntTargetSACViolationCIHigh}{4.93}
\newcommand{\FinalAntCandidateOverCoreReturnDelta}{2723.2}
\newcommand{\FinalAntCandidateOverCoreReturnCILow}{1502.9}
\newcommand{\FinalAntCandidateOverCoreReturnCIHigh}{4056.3}
\newcommand{\FinalAntCandidateOverCoreViolationDelta}{0.01}
\newcommand{\FinalAntCandidateOverCoreViolationCILow}{-1.09}
\newcommand{\FinalAntCandidateOverCoreViolationCIHigh}{1.10}
\newcommand{\FinalAntCandidateOverDirectMBPOReturnDelta}{73.4}
\newcommand{\FinalAntCandidateOverDirectMBPOReturnCILow}{-1947.6}
\newcommand{\FinalAntCandidateOverDirectMBPOReturnCIHigh}{1569.0}
\newcommand{\FinalAntCandidateOverDirectMBPOViolationDelta}{0.13}
\newcommand{\FinalAntCandidateOverDirectMBPOViolationCILow}{-1.02}
\newcommand{\FinalAntCandidateOverDirectMBPOViolationCIHigh}{1.04}
\newcommand{\FinalAntCandidateOverTargetSACReturnDelta}{1925.7}
\newcommand{\FinalAntCandidateOverTargetSACReturnCILow}{715.6}
\newcommand{\FinalAntCandidateOverTargetSACReturnCIHigh}{3322.3}
\newcommand{\FinalAntCandidateOverTargetSACViolationDelta}{-2.56}
\newcommand{\FinalAntCandidateOverTargetSACViolationCILow}{-3.74}
\newcommand{\FinalAntCandidateOverTargetSACViolationCIHigh}{-1.45}
\newcommand{\FinalAntCandidateOverDARCReturnDelta}{2222.7}
\newcommand{\FinalAntCandidateOverDARCReturnCILow}{931.7}
\newcommand{\FinalAntCandidateOverDARCReturnCIHigh}{3603.4}
\newcommand{\FinalAntCandidateOverDARCViolationDelta}{-0.42}
\newcommand{\FinalAntCandidateOverDARCViolationCILow}{-1.49}
\newcommand{\FinalAntCandidateOverDARCViolationCIHigh}{0.54}
\newcommand{\FinalAntCandidateOverRawSimReturnDelta}{2496.3}
\newcommand{\FinalAntCandidateOverRawSimReturnCILow}{1245.7}
\newcommand{\FinalAntCandidateOverRawSimReturnCIHigh}{3894.1}
\newcommand{\FinalAntCandidateOverRawSimViolationDelta}{0.98}
\newcommand{\FinalAntCandidateOverRawSimViolationCILow}{0.38}
\newcommand{\FinalAntCandidateOverRawSimViolationCIHigh}{1.62}
\newcommand{\FinalGravityDARCReturnMean}{1268.6}
\newcommand{\FinalGravityDARCReturnCILow}{948.8}
\newcommand{\FinalGravityDARCReturnCIHigh}{1570.3}
\newcommand{\FinalGravityDARCViolationMean}{44.23}
\newcommand{\FinalGravityDARCViolationCILow}{30.04}
\newcommand{\FinalGravityDARCViolationCIHigh}{60.58}
\newcommand{\FinalGravityDirectMBPOReturnMean}{4827.5}
\newcommand{\FinalGravityDirectMBPOReturnCILow}{4144.8}
\newcommand{\FinalGravityDirectMBPOReturnCIHigh}{5335.7}
\newcommand{\FinalGravityDirectMBPOViolationMean}{0.66}
\newcommand{\FinalGravityDirectMBPOViolationCILow}{0.19}
\newcommand{\FinalGravityDirectMBPOViolationCIHigh}{1.23}
\newcommand{\FinalGravityCandidateReturnMean}{4954.8}
\newcommand{\FinalGravityCandidateReturnCILow}{4675.7}
\newcommand{\FinalGravityCandidateReturnCIHigh}{5248.0}
\newcommand{\FinalGravityCandidateViolationMean}{0.20}
\newcommand{\FinalGravityCandidateViolationCILow}{0.03}
\newcommand{\FinalGravityCandidateViolationCIHigh}{0.47}
\newcommand{\FinalGravityCoreReturnMean}{4609.9}
\newcommand{\FinalGravityCoreReturnCILow}{4256.9}
\newcommand{\FinalGravityCoreReturnCIHigh}{4967.6}
\newcommand{\FinalGravityCoreViolationMean}{2.48}
\newcommand{\FinalGravityCoreViolationCILow}{0.16}
\newcommand{\FinalGravityCoreViolationCIHigh}{5.91}
\newcommand{\FinalGravityRawSimReturnMean}{1088.8}
\newcommand{\FinalGravityRawSimReturnCILow}{604.1}
\newcommand{\FinalGravityRawSimReturnCIHigh}{1572.6}
\newcommand{\FinalGravityRawSimViolationMean}{9.86}
\newcommand{\FinalGravityRawSimViolationCILow}{6.63}
\newcommand{\FinalGravityRawSimViolationCIHigh}{12.71}
\newcommand{\FinalGravityTargetSACReturnMean}{3770.3}
\newcommand{\FinalGravityTargetSACReturnCILow}{3388.9}
\newcommand{\FinalGravityTargetSACReturnCIHigh}{4143.1}
\newcommand{\FinalGravityTargetSACViolationMean}{9.52}
\newcommand{\FinalGravityTargetSACViolationCILow}{3.93}
\newcommand{\FinalGravityTargetSACViolationCIHigh}{15.93}
\newcommand{\FinalGravityCandidateOverCoreReturnDelta}{344.9}
\newcommand{\FinalGravityCandidateOverCoreReturnCILow}{-130.4}
\newcommand{\FinalGravityCandidateOverCoreReturnCIHigh}{805.4}
\newcommand{\FinalGravityCandidateOverCoreViolationDelta}{-2.29}
\newcommand{\FinalGravityCandidateOverCoreViolationCILow}{-5.84}
\newcommand{\FinalGravityCandidateOverCoreViolationCIHigh}{0.01}
\newcommand{\FinalGravityCandidateOverDirectMBPOReturnDelta}{127.3}
\newcommand{\FinalGravityCandidateOverDirectMBPOReturnCILow}{-398.3}
\newcommand{\FinalGravityCandidateOverDirectMBPOReturnCIHigh}{731.5}
\newcommand{\FinalGravityCandidateOverDirectMBPOViolationDelta}{-0.45}
\newcommand{\FinalGravityCandidateOverDirectMBPOViolationCILow}{-1.06}
\newcommand{\FinalGravityCandidateOverDirectMBPOViolationCIHigh}{0.12}
\newcommand{\FinalGravityCandidateOverTargetSACReturnDelta}{1184.5}
\newcommand{\FinalGravityCandidateOverTargetSACReturnCILow}{724.5}
\newcommand{\FinalGravityCandidateOverTargetSACReturnCIHigh}{1752.7}
\newcommand{\FinalGravityCandidateOverTargetSACViolationDelta}{-9.32}
\newcommand{\FinalGravityCandidateOverTargetSACViolationCILow}{-15.79}
\newcommand{\FinalGravityCandidateOverTargetSACViolationCIHigh}{-3.71}
\newcommand{\FinalGravityCandidateOverDARCReturnDelta}{3686.2}
\newcommand{\FinalGravityCandidateOverDARCReturnCILow}{3181.7}
\newcommand{\FinalGravityCandidateOverDARCReturnCIHigh}{4179.9}
\newcommand{\FinalGravityCandidateOverDARCViolationDelta}{-44.03}
\newcommand{\FinalGravityCandidateOverDARCViolationCILow}{-60.12}
\newcommand{\FinalGravityCandidateOverDARCViolationCIHigh}{-29.74}
\newcommand{\FinalGravityCandidateOverRawSimReturnDelta}{3866.0}
\newcommand{\FinalGravityCandidateOverRawSimReturnCILow}{3396.7}
\newcommand{\FinalGravityCandidateOverRawSimReturnCIHigh}{4348.4}
\newcommand{\FinalGravityCandidateOverRawSimViolationDelta}{-9.66}
\newcommand{\FinalGravityCandidateOverRawSimViolationCILow}{-12.56}
\newcommand{\FinalGravityCandidateOverRawSimViolationCIHigh}{-6.42}
\newcommand{\FinalWalkerDARCReturnMean}{252.5}
\newcommand{\FinalWalkerDARCReturnCILow}{168.0}
\newcommand{\FinalWalkerDARCReturnCIHigh}{330.8}
\newcommand{\FinalWalkerDARCViolationMean}{17.26}
\newcommand{\FinalWalkerDARCViolationCILow}{7.33}
\newcommand{\FinalWalkerDARCViolationCIHigh}{29.63}
\newcommand{\FinalWalkerDirectMBPOReturnMean}{350.7}
\newcommand{\FinalWalkerDirectMBPOReturnCILow}{291.4}
\newcommand{\FinalWalkerDirectMBPOReturnCIHigh}{408.5}
\newcommand{\FinalWalkerDirectMBPOViolationMean}{23.59}
\newcommand{\FinalWalkerDirectMBPOViolationCILow}{11.66}
\newcommand{\FinalWalkerDirectMBPOViolationCIHigh}{36.64}
\newcommand{\FinalWalkerCandidateReturnMean}{394.6}
\newcommand{\FinalWalkerCandidateReturnCILow}{353.8}
\newcommand{\FinalWalkerCandidateReturnCIHigh}{434.1}
\newcommand{\FinalWalkerCandidateViolationMean}{0.31}
\newcommand{\FinalWalkerCandidateViolationCILow}{0.27}
\newcommand{\FinalWalkerCandidateViolationCIHigh}{0.37}
\newcommand{\FinalWalkerCoreReturnMean}{434.4}
\newcommand{\FinalWalkerCoreReturnCILow}{310.4}
\newcommand{\FinalWalkerCoreReturnCIHigh}{600.9}
\newcommand{\FinalWalkerCoreViolationMean}{24.98}
\newcommand{\FinalWalkerCoreViolationCILow}{0.68}
\newcommand{\FinalWalkerCoreViolationCIHigh}{64.49}
\newcommand{\FinalWalkerRawSimReturnMean}{408.8}
\newcommand{\FinalWalkerRawSimReturnCILow}{265.6}
\newcommand{\FinalWalkerRawSimReturnCIHigh}{590.7}
\newcommand{\FinalWalkerRawSimViolationMean}{33.12}
\newcommand{\FinalWalkerRawSimViolationCILow}{5.26}
\newcommand{\FinalWalkerRawSimViolationCIHigh}{78.31}
\newcommand{\FinalWalkerTargetSACReturnMean}{261.2}
\newcommand{\FinalWalkerTargetSACReturnCILow}{204.8}
\newcommand{\FinalWalkerTargetSACReturnCIHigh}{320.4}
\newcommand{\FinalWalkerTargetSACViolationMean}{7.16}
\newcommand{\FinalWalkerTargetSACViolationCILow}{2.55}
\newcommand{\FinalWalkerTargetSACViolationCIHigh}{12.17}
\newcommand{\FinalWalkerCandidateOverCoreReturnDelta}{-39.8}
\newcommand{\FinalWalkerCandidateOverCoreReturnCILow}{-194.8}
\newcommand{\FinalWalkerCandidateOverCoreReturnCIHigh}{78.8}
\newcommand{\FinalWalkerCandidateOverCoreViolationDelta}{-24.67}
\newcommand{\FinalWalkerCandidateOverCoreViolationCILow}{-64.28}
\newcommand{\FinalWalkerCandidateOverCoreViolationCIHigh}{-0.53}
\newcommand{\FinalWalkerCandidateOverDirectMBPOReturnDelta}{43.9}
\newcommand{\FinalWalkerCandidateOverDirectMBPOReturnCILow}{-32.2}
\newcommand{\FinalWalkerCandidateOverDirectMBPOReturnCIHigh}{120.4}
\newcommand{\FinalWalkerCandidateOverDirectMBPOViolationDelta}{-23.29}
\newcommand{\FinalWalkerCandidateOverDirectMBPOViolationCILow}{-36.18}
\newcommand{\FinalWalkerCandidateOverDirectMBPOViolationCIHigh}{-11.21}
\newcommand{\FinalWalkerCandidateOverTargetSACReturnDelta}{133.4}
\newcommand{\FinalWalkerCandidateOverTargetSACReturnCILow}{67.6}
\newcommand{\FinalWalkerCandidateOverTargetSACReturnCIHigh}{197.9}
\newcommand{\FinalWalkerCandidateOverTargetSACViolationDelta}{-6.85}
\newcommand{\FinalWalkerCandidateOverTargetSACViolationCILow}{-11.84}
\newcommand{\FinalWalkerCandidateOverTargetSACViolationCIHigh}{-2.25}
\newcommand{\FinalWalkerCandidateOverDARCReturnDelta}{142.1}
\newcommand{\FinalWalkerCandidateOverDARCReturnCILow}{55.9}
\newcommand{\FinalWalkerCandidateOverDARCReturnCIHigh}{224.7}
\newcommand{\FinalWalkerCandidateOverDARCViolationDelta}{-16.95}
\newcommand{\FinalWalkerCandidateOverDARCViolationCILow}{-29.19}
\newcommand{\FinalWalkerCandidateOverDARCViolationCIHigh}{-7.20}
\newcommand{\FinalWalkerCandidateOverRawSimReturnDelta}{-14.3}
\newcommand{\FinalWalkerCandidateOverRawSimReturnCILow}{-188.4}
\newcommand{\FinalWalkerCandidateOverRawSimReturnCIHigh}{142.1}
\newcommand{\FinalWalkerCandidateOverRawSimViolationDelta}{-32.81}
\newcommand{\FinalWalkerCandidateOverRawSimViolationCILow}{-77.94}
\newcommand{\FinalWalkerCandidateOverRawSimViolationCIHigh}{-4.82}

%% file: generated/final_exact_routing/exact_routing_results_macros.tex
\newcommand{\ExactRoutingAntResidualCount}{0}
\newcommand{\ExactRoutingAntDirectCount}{10}
\newcommand{\ExactRoutingAntExactMatchCount}{10}
\newcommand{\ExactRoutingAntReturnDelta}{1529.9}
\newcommand{\ExactRoutingAntReturnCILow}{426.9}
\newcommand{\ExactRoutingAntReturnCIHigh}{2833.1}
\newcommand{\ExactRoutingAntViolationDelta}{0.72}
\newcommand{\ExactRoutingAntViolationCILow}{-0.05}
\newcommand{\ExactRoutingAntViolationCIHigh}{1.60}
\newcommand{\ExactRoutingAntRiskMarginDelta}{0.835}
\newcommand{\ExactRoutingAntRiskMarginCILow}{0.694}
\newcommand{\ExactRoutingAntRiskMarginCIHigh}{0.996}
\newcommand{\ExactRoutingGravityResidualCount}{0}
\newcommand{\ExactRoutingGravityDirectCount}{10}
\newcommand{\ExactRoutingGravityExactMatchCount}{10}
\newcommand{\ExactRoutingGravityReturnDelta}{188.0}
\newcommand{\ExactRoutingGravityReturnCILow}{-178.7}
\newcommand{\ExactRoutingGravityReturnCIHigh}{591.7}
\newcommand{\ExactRoutingGravityViolationDelta}{0.08}
\newcommand{\ExactRoutingGravityViolationCILow}{-0.11}
\newcommand{\ExactRoutingGravityViolationCIHigh}{0.37}
\newcommand{\ExactRoutingGravityRiskMarginDelta}{0.097}
\newcommand{\ExactRoutingGravityRiskMarginCILow}{0.066}
\newcommand{\ExactRoutingGravityRiskMarginCIHigh}{0.146}
\newcommand{\ExactRoutingWalkerResidualCount}{10}
\newcommand{\ExactRoutingWalkerDirectCount}{0}
\newcommand{\ExactRoutingWalkerExactMatchCount}{10}
\newcommand{\ExactRoutingWalkerReturnDelta}{-21.2}
\newcommand{\ExactRoutingWalkerReturnCILow}{-56.0}
\newcommand{\ExactRoutingWalkerReturnCIHigh}{8.7}
\newcommand{\ExactRoutingWalkerViolationDelta}{-0.04}
\newcommand{\ExactRoutingWalkerViolationCILow}{-0.19}
\newcommand{\ExactRoutingWalkerViolationCIHigh}{0.06}
\newcommand{\ExactRoutingWalkerRiskMarginDelta}{0.086}
\newcommand{\ExactRoutingWalkerRiskMarginCILow}{0.083}
\newcommand{\ExactRoutingWalkerRiskMarginCIHigh}{0.089}

%% file: sections/introduction.tex
\section{Introduction}

Simulation can supply control data without exposing a target system to every
exploratory action. Dynamics randomization and rapid adaptation use this advantage
to train reinforcement learning (RL) policies that tolerate or identify dynamics
variation~\citep{peng2018,kumar2021rma}. MBRL
instead fits a predictive model and improves the policy with imagined
transitions. Under a simulator-to-target dynamics shift, however, either reusing
the simulator model or fitting a target model from limited data can fail. Policy
optimization can then exploit errors that were rare under the model-fitting
distribution.

Probabilistic ensembles represent model uncertainty~\citep{chua2018pets}, and
short branched rollouts limit compounding error~\citep{janner2019mbpo}, but neither
decides whether simulator correction or direct target modeling is better for a
particular shift. Model fitting must also account for the state-action
distribution induced by policy learning~\citep{ross2012agnostic}. This leaves
two coupled questions: which target-model family should be deployed, and how
strongly should each imagined transition influence policy learning?

We study these questions in MC-WM. MC-WM fits
two candidates from the same initial target data: a residual correction to a
frozen simulator ensemble and a direct target ensemble. A disjoint calibration
partition measures standardized joint state-and-reward error and fixes the
lower-risk family for the run. Direct-candidate fitting uses an isolated random
stream, so model-family selection cannot change downstream policy or replay
initialization merely by consuming random numbers. The deployed model then
produces one-step imagined transitions whose policy-critic losses are weighted
by a learned confidence signal and deterministic validity predicates. On the
residual route, the state correction combines a locked sparse component with a
masked neural remainder; the support, coefficients, calibration errors, routing
decision, filter behavior, and target access are exported for every run.

The paper makes three contributions. First, it defines calibration-risk routing
between residual and direct target world models using disjoint fit, selection,
and calibration partitions. Second, it provides an exact routing
counterfactual: automatic, forced-residual, and forced-direct arms fit identical
candidates under matched random streams, and the automatic arm must exactly
reproduce its selected forced arm. Third, it evaluates the complete method and
its limits through 540 unique reported run cells from 541 completed executions:
a fresh ten-seed main comparison on three controlled dynamics shifts and
morphologies, the routing counterfactual, a source-identical predecessor
mechanism matrix, and a matched sparse-versus-dense residual study. The evidence
is deliberately bounded to target MuJoCo
simulators. It does not imply physical sim-to-real transfer, stable symbolic
discovery across seeds, reduced total target access, or a safety guarantee.

%% file: sections/related_work.tex
\section{Related Work}

\paragraph{Adaptation across dynamics.}
Visual domain randomization trains perception networks across randomized
rendering conditions~\citep{tobin2017}; dynamics randomization trains control
policies across randomized physics~\citep{peng2018}. Rapid Motor Adaptation (RMA) infers an
environment representation online~\citep{kumar2021rma}. DARC estimates a
source-to-target reward correction
from state-action and transition domain classifiers
~\citep{eysenbach2021darc}. Hybrid Offline-and-Online RL
with dynamics gaps (H2O+) combines limited offline target data with online
simulator interaction
~\citep{niu2025h2oplus}. MC-WM differs by fitting explicit residual and direct
target world-model candidates, selecting between them on held-out transition
risk, and weighting model-generated policy updates. DARC is included under the
same target-training and held-out evaluation contract as the other main methods.

Residual world-model adaptation is the closest model class.
\citet{lanier2026redraw} adapt visual latent world models using offline,
reward-unlabelled target data and latent-state dynamics residuals. MC-WM uses
reward-labelled online target data and observation-space
increments, places a locked sparse component and masked neural remainder on the
residual route, and allows calibration risk to reject that route in favor of a
direct target ensemble. These are architectural and protocol differences; they
do not establish dominance across the distinct visual-control and state-control
settings.

\paragraph{World models and policy optimization.}
Probabilistic Ensembles with Trajectory Sampling (PETS) represents predictive
uncertainty with probabilistic dynamics ensembles~\citep{chua2018pets}.
MBPO limits model bias with short rollouts
branched from real data~\citep{janner2019mbpo}. Dreamer learns behaviors through
latent imagination~\citep{hafner2020dreamer}. Temporal Difference
Learning for Model Predictive Control (TD-MPC) combines latent dynamics with a
terminal value estimate for planning~\citep{hansen2022tdmpc}. MC-WM uses
one-step rollouts so that the experiment primarily tests target-model choice,
correction, and weighting rather than long-horizon compounding.

\paragraph{Structured residuals and constrained updates.}
Sparse Identification of Nonlinear Dynamics (SINDy) estimates sparse dynamics
from a prescribed feature library~\citep{brunton2016sindy}. MC-WM applies its
sparse-regression principle to target-minus-simulator state increments, but the
sparse branch is evaluated as a capacity and inspectability choice rather than
assumed to improve control. Policy learning is based on SAC
~\citep{haarnoja2018sac}. The project-specific three-critic pessimistic actor is
related to ensemble methods such as Randomized Ensembled Double Q-Learning
(REDQ), but is not a REDQ implementation~\citep{chen2021redq}. Constrained Policy
Optimization optimizes an expected-cost constraint~\citep{achiam2017cpo}, whereas Inverse Constrained
Reinforcement Learning infers constraints from demonstrations
~\citep{malik2021icrl}. MC-WM's predicates are narrower: they
screen imagined transitions and do not constrain the executed target policy.

%% file: sections/method.tex
\section{Method}

\subsection{Problem Setting}

We use observed-state Markov decision process (MDP) approximations for the
simulator and target,
$\mathcal{M}_{\ell}=(\mathcal{S},\mathcal{A},P_{\ell},\rho_{\ell},
\gamma,p_0)$ indexed by $\ell\in\{\mathrm{S},\mathrm{T}\}$. They share the
state space $\mathcal{S}$, action space $\mathcal{A}$, policy-return discount
$\gamma\in[0,1)$, and initial-state distribution $p_0$, but may differ in the
transition kernel $P_{\ell}$ and expected one-step reward $\rho_{\ell}$. The
target objective is
\begin{equation}
J_{\mathcal{M}_{\mathrm{T}}}(\pi_{\phi})
=\mathbb{E}_{\mathcal{M}_{\mathrm{T}},\pi_{\phi}}
\left[\sum_{t=0}^{\infty}\gamma^t r_t\right],
\label{eq:target-objective}
\end{equation}
where $\pi_{\phi}$ is the policy with parameters $\phi$ and $r_t$ is the
observed target reward, with zero rewards after episode termination.
Equation~\eqref{eq:target-objective} defines the discounted task objective;
training uses the entropy-regularized, violation-aware surrogate in
Section~\ref{sec:policy-optimization}, and evaluation reports undiscounted
episode return. Observed state, action, and termination at time $t$ are
$s_t\in\mathbb{R}^{n_s}$, $a_t\in\mathbb{R}^{n_a}$, and
$\kappa_t\in\{0,1\}$; $x_t=[s_t;a_t]$ is the state-action input. The simulator
dataset is $\mathcal{D}_{\mathrm{S}}$, and the initial target dataset is
$\mathcal{D}_{\mathrm{T}}^{0}$. Appendix~\ref{app:notation} gives the complete
one-symbol--one-meaning ledger.

\subsection{Pipeline and Calibration-Risk Routing}
\label{sec:routing}

MC-WM fits two target-model candidates before
online policy learning (Figure~\ref{fig:pipeline}; Algorithm~\ref{alg:mcwm}
in Appendix~\ref{app:policy-details}). The residual candidate
$\widehat M_{\mathrm{R}}$ corrects a frozen simulator ensemble
$\widehat M_{\mathrm{S}}$, whereas the direct candidate
$\widehat M_{\mathrm{D}}$ fits the same probabilistic ensemble architecture
directly to target transitions. An untouched calibration split selects the
candidate with lower standardized one-step state-and-reward error. This route
is fixed for the run. The deployed model then supplies confidence-weighted
one-step transitions for target-domain SAC updates.

\begin{figure*}[t]
\centering
\resizebox{\textwidth}{!}{%
\begin{tikzpicture}[
  node distance=5mm and 5mm,
  box/.style={draw,rounded corners=1.5pt,align=center,minimum height=8mm,
              text width=21mm,fill=blue!4,font=\scriptsize},
  gate/.style={draw,rounded corners=1.5pt,align=center,minimum height=8mm,
              text width=21mm,fill=orange!8,font=\scriptsize},
  flow/.style={-Latex,semithick}]
\node[box] (sim) {Simulator data $\mathcal{D}_{\mathrm{S}}$};
\node[box,right=of sim] (base) {Fit and freeze $\widehat M_{\mathrm{S}}$};
\node[box,right=of base] (res) {Residual candidate\\$\widehat M_{\mathrm{R}}$};
\node[gate,right=of res] (route) {Calibration-risk route};
\node[gate,right=of route] (weight) {Confidence $w_i^{\Delta}$ + predicate $\chi_i$};
\node[box,right=of weight] (policy) {Weighted replay + SAC};
\node[box,below=of sim] (target) {Target data $\mathcal{D}_{\mathrm{T}}^{0}$};
\node[box,below=of res] (direct) {Direct candidate\\$\widehat M_{\mathrm{D}}$};
\coordinate (feedback) at ([yshift=-6mm]target.south);
\draw[flow] (sim) -- (base);
\draw[flow] (base) -- (res);
\draw[flow] (target) -- (direct);
\draw[flow] (target.north) -- ++(0,3mm) -| (res.south);
\draw[flow] (res) -- (route);
\draw[flow] (direct.east) -| (route.south);
\draw[flow] (route) -- (weight);
\draw[flow] (weight) -- (policy);
\draw[flow] (policy.south) |- (feedback) -- (target.south);
\end{tikzpicture}
}
\caption{MC-WM execution path. Initial target data fit a residual and a direct
candidate; a disjoint calibration partition fixes the deployed family. Online
target replay refits only the deployed state-and-reward model. Trust weights
scale losses rather than physical rewards; a separate violation cost is
described in Section~\ref{sec:policy-optimization}.}
\label{fig:pipeline}
\end{figure*}
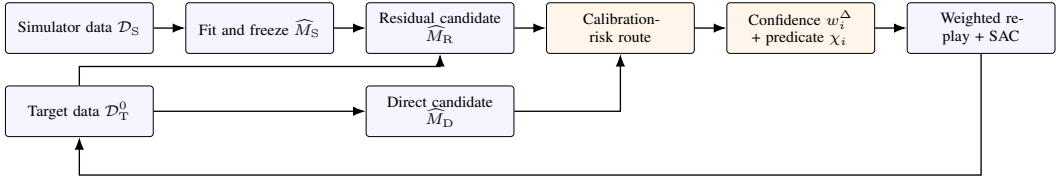

The initial 10,000 target transitions are split reproducibly into index sets
$\mathcal{I}_{\mathrm{fit}}$, $\mathcal{I}_{\mathrm{sel}}$, and
$\mathcal{I}_{\mathrm{cal}}$ containing 80\%, 10\%, and 10\% of the rows. The
fit split estimates parameters; the selection split chooses sparse features
and supplies early stopping for the direct ensemble; neither candidate uses
the calibration split during fitting or early stopping. Define the joint
one-step target
\begin{equation}
z_i=[s_{i+1}-s_i;r_i]\in\mathbb{R}^{n_s+1}.
\label{eq:routing-target}
\end{equation}
For a dataset sample $i$, $s_{i+1}$ denotes its observed successor, not the
next row of a shuffled dataset.
For candidate $\mathfrak m\in\{\mathrm{R},\mathrm{D}\}$, let
$\widehat z_i^{(\mathfrak m)}$ be its deterministic ensemble-mean prediction
and let $\varsigma_k$ be the standard deviation of coordinate $k$ of $z_i$ on
$\mathcal{I}_{\mathrm{fit}}$, replaced by one when it is at most $10^{-6}$.
The registered routing risk is
\begin{equation}
\mathcal{R}_{\mathfrak m}=
\left[
\frac{1}{|\mathcal{I}_{\mathrm{cal}}|(n_s+1)}
\sum_{i\in\mathcal{I}_{\mathrm{cal}}}
\sum_{k=1}^{n_s+1}
\left(
\frac{\widehat z_{i,k}^{(\mathfrak m)}-z_{i,k}}{\varsigma_k}
\right)^2
\right]^{1/2}.
\label{eq:routing-risk}
\end{equation}
Equation~\eqref{eq:routing-risk} evaluates the target in
Equation~\eqref{eq:routing-target} after coordinate standardization. MC-WM
deploys
\begin{equation}
\mathfrak m^{\star}=
\begin{cases}
\mathrm{D}, & \mathcal{R}_{\mathrm{D}}<\mathcal{R}_{\mathrm{R}},\\
\mathrm{R}, & \text{otherwise},
\end{cases}
\qquad
\widehat M_{\mathrm{T}}=\widehat M_{\mathfrak m^{\star}}.
\label{eq:model-family-route}
\end{equation}
Thus a tie retains the residual candidate. After Equation~\eqref{eq:model-family-route}
fixes the route, all initial target rows enter online replay. No validation
episode or test trajectory participates in this model-family decision.

Each probabilistic ensemble contains $K=5$ multilayer perceptron (MLP) members.
Every MLP has three 200-unit sigmoid linear unit (SiLU) hidden layers and predicts
a Gaussian mean and diagonal variance for state increment and reward. Input
and target normalizers are fixed at the candidate's initial fit. On the direct
route, $\widehat M_{\mathrm{D}}$ is warm-start refit every 1,000 online target
steps on at most 50,000 replay transitions. On the residual route, the
correction described next is refit at the same frequency and replay cap while
$\widehat M_{\mathrm{S}}$ remains frozen.

Direct-candidate fitting uses a dedicated random stream derived from the seed,
preserving subsequent policy, replay, and optimizer initialization. Automatic
and forced-route arms fit both candidates in the same order and intervene only
on deployment in Equation~\eqref{eq:model-family-route}. The automatic arm must
exactly reproduce its selected forced arm; Appendix~\ref{app:exact-routing}
details this isolation and matching test.

\subsection{Locked-Support Residual Candidate}

Let $\widehat u_{\mathrm{S}}(x)$ and
$\widehat\rho_{\mathrm{S}}(x)$ denote the frozen simulator ensemble's mean
state-increment and reward predictions. For target transition $i$, the state
correction target is
\begin{equation}
y_i^{\Delta s}=s_{i+1}-s_i-\widehat u_{\mathrm{S}}(x_i).
\label{eq:state-correction-target}
\end{equation}
Equation~\eqref{eq:state-correction-target} removes the simulator-predicted
increment from the observed target increment.
The residual candidate predicts
\begin{align}
\widehat s_{t+1}^{(\mathrm{R})}
&=s_t+\widehat u_{\mathrm{S}}(x_t)+\widehat\Delta_s(x_t),
\label{eq:corrected-state}\\
\widehat r_t^{(\mathrm{R})}
&=\widehat\rho_{\mathrm{S}}(x_t)+\widehat\Delta_{r,\zeta}(x_t),
\label{eq:corrected-reward}
\end{align}
Equations~\eqref{eq:corrected-state} and \eqref{eq:corrected-reward} define the
residual candidate's state and reward predictions, respectively,
where $\widehat\Delta_{r,\zeta}$ is a one-hidden-layer, 64-unit rectified
linear unit (ReLU) reward
correction with parameters $\zeta$. The additive state correction is
\begin{equation}
\widehat\Delta_s(x)
=B\widetilde\Theta(x)
+g_{\psi}\!\left(\mu\odot\widetilde\Theta(x)\right).
\label{eq:additive-correction}
\end{equation}
Equation~\eqref{eq:additive-correction} defines the complete state correction.
Here, $\mathcal{F}$ is the candidate feature set with a fixed evaluation order,
$\Theta(x)$ evaluates its features as a vector,
$\widetilde\Theta(x)\in\mathbb{R}^{p}$ is its root-mean-square-scaled deployed
subset, $B\in\mathbb{R}^{n_s\times p}$ is the sparse coefficient matrix,
$\mu\in\{0,1\}^{p}$ marks features active in at least one state coordinate,
$g_{\psi}$ is the masked neural remainder, and $\odot$ is elementwise
multiplication. The remainder has one 32-unit SiLU hidden layer and a
zero-initialized output layer. It therefore starts at zero and cannot consume
a feature outside the reported support.

A degree-two polynomial library supplies the base features. Sequentially
thresholded least squares (STLSQ) uses fixed ridge and threshold values;
deterministic additions must pass a selection-partition mean squared error
(MSE) gate and a coefficient screen. The accepted feature order and support are
then locked. Online residual refits update coefficients, the masked neural
remainder, reward, and eligible termination heads without adding features, and
weight rows by proximity to recent policy states. Appendix~\ref{app:residual-details}
gives the selection gate in Equation~\eqref{eq:feature-gate}, the refit weight
in Equation~\eqref{eq:policy-weight}, and the termination prediction in
Equation~\eqref{eq:corrected-termination}.

\subsection{Confidence and Predicate Weights}

For target sample $i$, define the deployed-model state error
$e_i=n_s^{-1}\lVert\widehat s_{i+1}-s_{i+1}\rVert_2^2$. Let $\tau_e$ be an
positive exponential moving average of the batch median error, initialized at
$10^{-3}$ and updated with weight 0.1 on the new median (floored at $10^{-8}$), and
$c_i=(1+e_i/\tau_e)^{-1}$. MC-WM fits a nonpositive log-confidence critic
$Q^{\Delta}_{\xi}$. Its ensemble signature
$\boldsymbol{\sigma}_{\mathrm{ens}}(s,a)\in
\mathbb{R}^{n_s+1}_{\geq0}$ is the coordinatewise standard deviation of the
direct target ensemble on the direct route and of the frozen simulator
ensemble on the residual route. The regression target is
\begin{equation}
q_i^{\mathrm{tgt}}=\log\max(c_i,10^{-8})+
\gamma_{\Delta}(1-\kappa_i)
Q^{\Delta}_{\bar\xi}\!\left(
s_{i+1},a_{i+1}^{\mathrm{det}},
\boldsymbol{\sigma}_{\mathrm{ens}}
(s_{i+1},a_{i+1}^{\mathrm{det}})
\right),
\label{eq:confidence-target}
\end{equation}
where $a_{i+1}^{\mathrm{det}}$ is the current policy's deterministic action
(the squashed Gaussian mean) at $s_{i+1}$, $\xi$ is the confidence-critic
parameter vector, $\bar\xi$ denotes its Polyak-averaged target parameters, and
$\gamma_{\Delta}=0.5$ is distinct from the policy discount. The imagined
transition weight is
\begin{equation}
w_i^{\Delta}=\exp\!\left(Q^{\Delta}_{\xi}\!\left(
s_i,a_i,\boldsymbol{\sigma}_{\mathrm{ens}}(s_i,a_i)
\right)\right)\in(0,1].
\label{eq:confidence-weight}
\end{equation}
Equation~\eqref{eq:confidence-target} trains the critic, while
Equation~\eqref{eq:confidence-weight} converts its nonpositive output to a
weight. This value is a learned ranking signal, not a calibrated probability.

The deterministic predicate set $\mathcal{C}$ contains broad
morphology-specific representation, transition-magnitude, and reward-range
checks. The Walker set additionally rejects non-finite predictions. The
HalfCheetah and Walker experiments use nonempty sets; the Ant environment uses
an empty set. For imagined transition $i$, the indicator $\chi_i$ is one only
when every configured predicate passes. The final trust weight is
\begin{equation}
w_i=w_i^{\Delta}\chi_i,
\label{eq:combined-weight}
\end{equation}
and rows with $w_i\leq0.01$ are discarded. Predicates remove known invalid
predictions but do not constrain target actions.

\subsection{Violation-Aware Policy Optimization and Checkpoint Selection}
\label{sec:policy-optimization}

Every 250 target steps, the current policy acts from 400 target-replay states,
and $\widehat M_{\mathrm{T}}$ generates one-step transitions. Physical rewards
are stored without confidence scaling. A separate binary violation indicator
$\iota_i\in\{0,1\}$ is supplied by target-environment diagnostics for real
transitions and by the same environment's state predicate for imagined
transitions. At online step $t\in\{1,\ldots,N_{\mathrm{on}}\}$, replay sampling
uses
\begin{equation}
\lambda_{\mathrm{v}}(t)=10\max\!\left(
0,\frac{t-0.5N_{\mathrm{on}}}{0.5N_{\mathrm{on}}}
\right),
\qquad
r_i^{\mathrm{train}}=r_i-\lambda_{\mathrm{v}}(t)\iota_i.
\label{eq:violation-cost}
\end{equation}
Thus Equation~\eqref{eq:violation-cost} leaves the violation cost at zero
through the first half of online training and raises it linearly to 10; the
cost is distinct from the trust weight in
Equation~\eqref{eq:combined-weight}.

Let $\mathcal{I}$ index a minibatch of size $b=256$, let $Q_{\eta_j}$ be policy
critic $j\in\{1,2,3\}$ with parameters $\eta_j$, and let $y_i^Q$ be the
SAC target formed with scale-normalized $r_i^{\mathrm{train}}$, the minimum of
the three target critics, and the entropy-adjusted stochastic next action.
The critic loss is
\begin{equation}
\mathcal{L}_Q=
\frac{1}{3b}\sum_{j=1}^{3}\sum_{i\in\mathcal{I}}
w_i\left(Q_{\eta_j}(s_i,a_i)-y_i^Q\right)^2
+\lambda_{\mathrm{ens}}\frac{1}{b}\sum_{i\in\mathcal{I}}
\operatorname{Std}_{j}\!\left[Q_{\eta_j}(s_i,a_i)\right],
\label{eq:critic-loss}
\end{equation}
In Equation~\eqref{eq:critic-loss}, $w_i=1$ for target transitions,
$\lambda_{\mathrm{ens}}=0.01$, and $\operatorname{Std}_{j}$ takes the sample
standard deviation across the three policy critics. The actor maximizes the critic mean minus two
critic standard deviations plus the SAC entropy term, with a
$10^{-3}$ behavior-cloning MSE penalty. Policy learners use scale-only running
reward normalization.

An initial simulator-to-residual state root mean squared error (RMSE) ratio
$\varrho_{\mathrm{cal}}$ on
$\mathcal{I}_{\mathrm{cal}}$ fixes whether actor updates mix target and imagined
states; this actor route is separate from Equation~\eqref{eq:model-family-route}.
Fixed validation checkpoints may restore an eligible actor before the one-time
held-out test, so validation is counted as model-selection access. The exact
actor band, restoration inequalities, and complete training procedure appear
in Appendix~\ref{app:policy-details}.

%% file: sections/experiments.tex
\section{Experimental Protocol}
\label{sec:experiments}

\subsection{Controlled Target Domains}

Experiments use the MuJoCo physics engine~\citep{todorov2012mujoco} through
Gymnasium 1.2.0 \texttt{v5} tasks. Each simulator--target pair preserves the
task morphology and observation and action dimensions while changing one
controlled dynamics factor. The final-confirmation matrix contains three
environments, spanning three morphologies and three shift types
(Table~\ref{tab:environments}). Appendix~\ref{app:environments} gives the exact
physics, reward, and termination transformations. We call the held-out member
of each pair the \emph{target}; these are simulator-to-simulator experiments,
not physical sim-to-real deployments.

The task coordinates are $q_z$, HalfCheetah's root vertical generalized
coordinate at observation index 0; $v_x$, Ant's forward velocity at index 13;
and $s^{(0)}$, Walker's root-height observation. The HalfCheetah coordinate is
zero-centered, not absolute torso height. Gravity is in $\mathrm{m\,s^{-2}}$.

\begin{table}[H]
\caption{Controlled simulator-to-target environments. Each reported
constraint is identical in the simulator and target member of a pair.}
\label{tab:environments}
\centering
\scriptsize
\resizebox{\linewidth}{!}{%
\begin{tabular}{llll}
\toprule
Name & Base task & Simulator $\rightarrow$ target dynamics & Counted violation \\
\midrule
Ant actuator & Ant-v5 & 8 active $\rightarrow$ actions 4--7 disabled & $|v_x|>1.5$ or rear-wall hit \\
Gravity & HalfCheetah-v5 & vertical gravity $-19.62\rightarrow-9.81$ & $q_z>0.2$ \\
Friction Walker & Walker2d-v5 & friction factor $0.3\rightarrow1.0$ & $s^{(0)}>1.25$ \\
\bottomrule
\end{tabular}}
\end{table}

The Gravity and Friction
Walker thresholds apply continuous quadratic reward penalties, whereas the Ant
environment combines a quadratic velocity penalty with termination when the
world-frame base position is below $-3$ along the first horizontal coordinate.
A violation is counted at each penalized step and at a wall
termination. Counts are therefore comparable only within an environment and
are never pooled across constraint definitions.

The Ant construction is inspired by the Ant-Broken constraint-transfer setting
of \citet{malik2021icrl}, but it is not a reproduction. It uses a
27-dimensional Gymnasium-v5 observation, four disabled actuator channels, a
project-specific radial reward, and a shared velocity cap. The deterministic
predicate set used by MC-WM does not contain an Ant-specific predicate in the
reported protocol; Ant violations remain environment outcomes and training
costs rather than predicate-filter events.

\subsection{Candidate Freezing and Main Comparators}

The final candidate was fixed before the confirmation endpoints were read. A
registered seven-seed development screen replaced the predecessor's indirect
state-error-gap route with the calibration risk in
Equation~\eqref{eq:routing-risk}. A subsequent implementation audit found that
fitting the direct candidate had consumed the global random stream
before policy initialization. We corrected this causal confound by isolating the
component random stream as described in Section~\ref{sec:routing}, froze the corrected source,
and discarded the affected confirmation endpoints from the submission evidence.
The final ten-seed bank was then sampled separately. It excludes all 40 seeds
registered by the preceding development and confirmation studies. Development
outcomes are used only for algorithm selection; all main performance claims use
the fresh, random-stream-isolated confirmation matrix.

The main matrix contains the frozen calibration-routed MC-WM candidate in
Algorithm~\ref{alg:mcwm} and five controls. \emph{Residual MC-WM core} retains
the sparse-plus-neural residual,
confidence critic, and policy-aware refitting, but fixes residual deployment
and removes adaptive update routing, the violation cost, and validation-based
checkpoint restoration. \emph{Direct target MBPO} fits the same probabilistic
world-model architecture directly to target transitions and uses one-step
rollouts. \emph{Target SAC} learns from target transitions without a learned
world model. Our matched-protocol \emph{DARC} implementation uses the published additive
state-action and state-action-next-state classifiers, 32-unit hidden layers,
unit Gaussian input noise, and starts reward correction after 4,000 online
updates~\citep{eysenbach2021darc}. The warmup is scaled to our 40,000-step
online phase and is not presented as a reproduction of DARC's original longer
schedule. \emph{Raw simulator SAC} learns for 50,000 interactions in the
uncorrected simulator and receives no target-domain training transition. All
SAC-labelled methods use the same project-specific policy backbone described
in Section~\ref{sec:policy-optimization}; ``SAC'' does not denote an otherwise
unchanged reference implementation.

The primary comparisons are candidate versus residual MC-WM core and direct
target MBPO. They test whether calibrated family selection improves over a
fixed residual and whether simulator correction improves over direct target
modeling. Target SAC, DARC, and raw simulator SAC are secondary controls for
model-free target learning, classifier-based adaptation, and zero-target-
training transfer, respectively. Every method is run for ten independently
trained seeds on all three environments, giving $3\times6\times10=180$
confirmation runs from one frozen source tree.

\subsection{Training and Evaluation Access}

Table~\ref{tab:budgets} separates interactions that can change model or policy
parameters from target validation and final testing. Each adaptive method
receives 10,000 initial target transitions and 40,000 online target
transitions. Candidate and residual runs also use 50,000 simulator transitions
to fit the frozen source model. DARC uses 10,000 initial and 40,000 online
simulator transitions. Direct target MBPO and target SAC use no simulator
training transition. Raw simulator SAC instead uses 50,000 simulator
transitions and zero target training transitions.

Whenever a method receives an initial simulator or target dataset, each action
coordinate is sampled independently and uniformly within the environment's
action bounds using a dataset-specific seed. Online loops index interactions
from $t=1$: actions are sampled from the same uniform distribution for
$t<2{,}000$ (steps 1--1,999), and from the stochastic policy beginning at
$t=2{,}000$. DARC applies this online cutoff simultaneously in its simulator
and target streams; raw simulator SAC applies it only in the simulator. These
random-action transitions are included in the training budgets in
Table~\ref{tab:budgets}.

Validation uses seeds 10,000--10,004 for five episodes at each of eight
checkpoints. Adaptive methods are evaluated every 5,000 online steps; raw
simulator SAC is evaluated every 6,250 simulator steps so that it has the same
number of validation calls over its 50,000-step budget. The one-shot final test
uses seeds 20,000--20,019 for 20 episodes. An episode has at most 1,000 steps,
so an adaptive run can access at most 110,000 target transitions: 50,000 for
training, 40,000 for validation, and 20,000 for final testing. Early
terminations reduce the realized count.

Validation data never enter gradient updates. They are nevertheless model-
selection access: the candidate and its exact routing variants can restore a
registered actor checkpoint from the validation trajectory. We therefore
report training, validation, test, and total target access separately and do
not describe the 50,000-transition training budget as total target access.
Figure~\ref{fig:target-ledger} in Appendix~\ref{app:target-access} reports
realized access by environment and method.
Learning curves use cumulative target training plus validation interactions on
the horizontal axis, which permits the registered method-specific checkpoint
grids without treating simulator steps as target access.

\begin{table}[H]
\caption{Registered per-seed environment access. Episode lengths determine
the realized validation and test transition counts, which are recorded in the
interaction ledger.}
\label{tab:budgets}
\centering
\small
\begin{tabular}{lrrrr}
\toprule
Method & Simulator train & Target train & Validation & Final test \\
\midrule
Calibration-routed MC-WM & 50,000 & 50,000 & $8\times5$ episodes & 20 episodes \\
Residual MC-WM core & 50,000 & 50,000 & $8\times5$ episodes & 20 episodes \\
Direct target MBPO / target SAC & 0 & 50,000 & $8\times5$ episodes & 20 episodes \\
DARC & 50,000 & 50,000 & $8\times5$ episodes & 20 episodes \\
Raw simulator SAC & 50,000 & 0 & $8\times5$ episodes & 20 episodes \\
\bottomrule
\end{tabular}
\end{table}

\subsection{Registered Counterfactual and Ablations}

The exact routing study reuses the 30 final-candidate cells and adds 60 forced
controls. All arms fit both candidates with equal non-routing arguments and
frozen pre-deployment diagnostics; the automatic arm must exactly reproduce the
metric stream and sanitized summary of its selected forced arm. The registered
contrast is automatic routing minus the unselected alternative. Separate
$3\times8\times10=240$ predecessor-mechanism and
$3\times2\times10=60$ fixed-route sparse-versus-dense matrices test component
and residual-backend claims. Appendices~\ref{app:exact-routing},
\ref{app:predecessor}, and \ref{app:sparse-dense} give the respective
interventions and matching rules.

\subsection{Endpoints, Diagnostics, and Statistics}

The statistical unit is one independently trained seed. For each environment
and method, we report undiscounted episode return and counted violations,
aggregated as the mean and sample standard deviation (SD) of seed-level
20-episode test means. Pre-registered paired comparisons subtract outcomes
within the same environment and training seed and use a 20,000-resample
percentile-bootstrap 95\% confidence interval (CI) for the cross-seed mean.
The main study uses candidate minus control. The routing study reports both
candidate minus each forced arm and, as its causal contrast, automatic routing
minus the forced alternative selected per seed. The mechanism matrix uses
ablation minus predecessor, and the sparse-versus-dense study uses dense minus
sparse. We label a direction only when its CI excludes zero, report every
favorable, unfavorable, and unresolved comparison, and make no unregistered
family-wise claim. Returns are not pooled across environments because their
reward scales differ.

Each final-candidate run additionally reports the selected world-model family,
both calibration risks, active and deployed sparse-support sizes, cross-seed
support Jaccard overlap, confidence weights, and predicate rejection rate. The
confidence diagnostic uses the first 1,000 stored target-replay transitions at
the final model fit, and rejection rates aggregate imagined-rollout checks. The
sparse-versus-dense study also reports calibration transition error and
trainable state-residual parameters for both backends. These diagnostics test
route consistency, quantify model fit and capacity, and distinguish inspection
of one fitted symbolic correction from stability of that correction across
seeds. No support-stability claim is inferred from equation export alone;
Appendix~\ref{app:auditability} gives the fixed display rule.

The four matrices contain 540 unique reported cells from 541 completed
executions. Aggregation rejects provenance, seed, budget, evaluation, or
artifact drift, and generates every table and figure from validated artifacts.
Appendix~\ref{app:reproducibility} gives the matrix ledger, replacement
execution, frozen identities, and generation command.

%% file: sections/results.tex
\section{Results}

\subsection{Held-Out Main Comparison}

Table~\ref{tab:main-results} reports the one-shot held-out outcomes for the six
methods. Figure~\ref{fig:learning-curves} in Appendix~\ref{app:target-access}
shows validation trajectories against cumulative target access. These
trajectories can trigger candidate checkpoint restoration; held-out test
episodes remain disjoint from that decision.

Relative to target SAC, calibration-routed MC-WM increases return on Ant by
$\FinalAntCandidateOverTargetSACReturnDelta$
[$\FinalAntCandidateOverTargetSACReturnCILow$,
$\FinalAntCandidateOverTargetSACReturnCIHigh$], on HalfCheetah gravity by
$\FinalGravityCandidateOverTargetSACReturnDelta$
[$\FinalGravityCandidateOverTargetSACReturnCILow$,
$\FinalGravityCandidateOverTargetSACReturnCIHigh$], and on Walker friction by
$\FinalWalkerCandidateOverTargetSACReturnDelta$
[$\FinalWalkerCandidateOverTargetSACReturnCILow$,
$\FinalWalkerCandidateOverTargetSACReturnCIHigh$]. The paired violation
differences are
$\FinalAntCandidateOverTargetSACViolationDelta$
[$\FinalAntCandidateOverTargetSACViolationCILow$,
$\FinalAntCandidateOverTargetSACViolationCIHigh$],
$\FinalGravityCandidateOverTargetSACViolationDelta$
[$\FinalGravityCandidateOverTargetSACViolationCILow$,
$\FinalGravityCandidateOverTargetSACViolationCIHigh$], and
$\FinalWalkerCandidateOverTargetSACViolationDelta$
[$\FinalWalkerCandidateOverTargetSACViolationCILow$,
$\FinalWalkerCandidateOverTargetSACViolationCIHigh$], respectively; all three
favor MC-WM. Candidate return also exceeds DARC in every environment. DARC
violation differences favor MC-WM on gravity and Walker and are unresolved on
Ant.

Direct target MBPO is the strongest model-based comparator. Its return interval
against the candidate includes zero in all three environments: Ant
$\FinalAntCandidateOverDirectMBPOReturnDelta$
[$\FinalAntCandidateOverDirectMBPOReturnCILow$,
$\FinalAntCandidateOverDirectMBPOReturnCIHigh$], gravity
$\FinalGravityCandidateOverDirectMBPOReturnDelta$
[$\FinalGravityCandidateOverDirectMBPOReturnCILow$,
$\FinalGravityCandidateOverDirectMBPOReturnCIHigh$], and Walker
$\FinalWalkerCandidateOverDirectMBPOReturnDelta$
[$\FinalWalkerCandidateOverDirectMBPOReturnCILow$,
$\FinalWalkerCandidateOverDirectMBPOReturnCIHigh$]. On Walker, however, MC-WM
reduces violations by
$\FinalWalkerCandidateOverDirectMBPOViolationDelta$
[$\FinalWalkerCandidateOverDirectMBPOViolationCILow$,
$\FinalWalkerCandidateOverDirectMBPOViolationCIHigh$]. Compared with fixed
residual MC-WM core, the candidate improves Ant return by
$\FinalAntCandidateOverCoreReturnDelta$
[$\FinalAntCandidateOverCoreReturnCILow$,
$\FinalAntCandidateOverCoreReturnCIHigh$]; gravity and Walker return differences
are unresolved, while Walker violations favor the candidate. Against raw
simulator SAC, the candidate improves Ant and gravity return but incurs more Ant
violations; Walker return is unresolved and Walker violations favor the
candidate. These contrasts rule out a claim of dominance on every endpoint.

\begin{table*}[t]
\caption{Held-out target-domain outcomes over ten independently trained seeds
(mean $\pm$ sample SD). Higher return and fewer violations are better. Test
seeds are fixed across methods; target training, validation, and test access are
reported separately in Appendix~\ref{app:target-access}.}
\label{tab:main-results}
\centering
\scriptsize
\resizebox{\textwidth}{!}{\input{generated/final_confirmation/algorithm_confirmation_endpoints.tex}}
\end{table*}

\subsection{Exact Routing Counterfactual}

The calibration criterion selects the direct target family in all ten Ant and
all ten gravity runs, and the residual family in all ten Walker runs
(Table~\ref{tab:exact-routing}). Every automatic run exactly matches its selected
forced arm, establishing that the override itself does not perturb the selected
training trajectory. The causal contrast therefore compares automatic routing
with the unselected forced alternative under matched random streams.

The automatic-minus-alternative return differences are
$\ExactRoutingAntReturnDelta$
[$\ExactRoutingAntReturnCILow$, $\ExactRoutingAntReturnCIHigh$] on Ant,
$\ExactRoutingGravityReturnDelta$
[$\ExactRoutingGravityReturnCILow$, $\ExactRoutingGravityReturnCIHigh$] on
gravity, and $\ExactRoutingWalkerReturnDelta$
[$\ExactRoutingWalkerReturnCILow$, $\ExactRoutingWalkerReturnCIHigh$] on Walker.
The corresponding violation differences are
$\ExactRoutingAntViolationDelta$
[$\ExactRoutingAntViolationCILow$, $\ExactRoutingAntViolationCIHigh$],
$\ExactRoutingGravityViolationDelta$
[$\ExactRoutingGravityViolationCILow$, $\ExactRoutingGravityViolationCIHigh$],
and $\ExactRoutingWalkerViolationDelta$
[$\ExactRoutingWalkerViolationCILow$, $\ExactRoutingWalkerViolationCIHigh$].
Ant supplies a resolved route-return benefit; gravity and Walker return effects
and all route-level violation effects are unresolved.

\begin{table*}[t]
\caption{Exact route intervention over ten paired seeds. Selected R/D is the
automatic residual/direct count; exact match compares the complete metric stream
with the selected forced arm. Differences are automatic routing minus the
unselected forced alternative, so positive return and negative violations favor
the selector.}
\label{tab:exact-routing}
\centering
\scriptsize
\resizebox{\textwidth}{!}{\input{generated/final_exact_routing/exact_routing_results.tex}}
\end{table*}

\subsection{Mechanisms, Capacity, and Inspectability}

The studies in Appendix~\ref{app:ablations} distinguish predecessor mechanisms
from the final selector. The predecessor matrix supports validation checkpoint restoration on Ant and
Walker, avoiding a fixed residual route on Ant, and a return--violation
tradeoff from the violation cost; separate confidence and predicate removals
remain unresolved. The fixed-route sparse comparison resolves no control
endpoint, but uses fewer state-residual parameters in every environment while
calibration risk favors different backends by environment. Auditability records
show low cross-seed support overlap despite per-run equation export. Appendix
Tables~\ref{tab:auditability}, \ref{tab:mechanism-ablations}, and
\ref{tab:sparse-dense} report the complete contrasts and diagnostics.
The exported sparse-coordinate corrections in
Equations~\eqref{eq:sparse-ant_wall_broken_soft_ceiling-example},
\eqref{eq:sparse-gravity_soft_ceiling-example}, and
\eqref{eq:sparse-friction_walker_soft_ceiling-example} make this per-run
inspectability concrete; they are not governing laws shared across seeds.

%% file: generated/final_confirmation/algorithm_confirmation_endpoints.tex
\begin{tabular}{lrrrrrr}
\toprule
 & \multicolumn{2}{c}{Ant actuator} & \multicolumn{2}{c}{HalfCheetah gravity} & \multicolumn{2}{c}{Walker friction} \\
\cmidrule(lr){2-3} \cmidrule(lr){4-5} \cmidrule(lr){6-7}
Method & Return $\uparrow$ & Violations $\downarrow$ & Return $\uparrow$ & Violations $\downarrow$ & Return $\uparrow$ & Violations $\downarrow$ \\
\midrule
\textbf{Calibration-routed MC-WM} & 4084.9 $\pm$ 2105.9 & 1.39 $\pm$ 1.19 & 4954.8 $\pm$ 488.6 & 0.20 $\pm$ 0.40 & 394.6 $\pm$ 68.4 & 0.31 $\pm$ 0.09 \\
MC-WM core & 1361.7 $\pm$ 527.6 & 1.38 $\pm$ 1.48 & 4609.9 $\pm$ 602.0 & 2.48 $\pm$ 5.31 & 434.4 $\pm$ 247.8 & 24.98 $\pm$ 59.24 \\
Direct target MBPO & 4011.5 $\pm$ 3346.2 & 1.26 $\pm$ 1.32 & 4827.5 $\pm$ 1043.2 & 0.66 $\pm$ 0.89 & 350.7 $\pm$ 100.7 & 23.59 $\pm$ 21.45 \\
Target SAC & 2159.2 $\pm$ 461.1 & 3.95 $\pm$ 1.63 & 3770.3 $\pm$ 644.4 & 9.52 $\pm$ 10.24 & 261.2 $\pm$ 98.8 & 7.16 $\pm$ 8.19 \\
DARC & 1862.2 $\pm$ 534.7 & 1.80 $\pm$ 2.38 & 1268.6 $\pm$ 528.1 & 44.23 $\pm$ 26.03 & 252.5 $\pm$ 138.0 & 17.26 $\pm$ 19.03 \\
Raw simulator SAC & 1588.6 $\pm$ 416.2 & 0.41 $\pm$ 0.35 & 1088.8 $\pm$ 834.0 & 9.86 $\pm$ 5.26 & 408.8 $\pm$ 279.9 & 33.12 $\pm$ 68.54 \\
\bottomrule
\end{tabular}

%% file: generated/final_exact_routing/exact_routing_results.tex
\begin{tabular}{lrrrr}
\toprule
Environment & Selected R/D & Exact match & Return difference (95\% CI) & Violation difference (95\% CI) \\
\midrule
Ant actuator & 0/10 & 10/10 & 1529.9 [426.9, 2833.1] & 0.72 [-0.05, 1.60] \\
HalfCheetah gravity & 0/10 & 10/10 & 188.0 [-178.7, 591.7] & 0.08 [-0.11, 0.37] \\
Walker friction & 10/0 & 10/10 & -21.2 [-56.0, 8.7] & -0.04 [-0.19, 0.06] \\
\bottomrule
\end{tabular}

%% file: sections/discussion.tex
\section{Discussion}

The strongest result is model-family choice rather than residual modeling
alone. Calibration risk routes Ant and gravity to direct target models and
Walker to residual models. Exact selected-arm equality rules out an independent
override effect; Ant shows a return benefit with no resolved route harm
elsewhere. The main matrix also favors MC-WM over target SAC and matched DARC,
but return relative to direct target MBPO remains unresolved. The supported
claim is therefore calibrated choice under these shifts, not universal routing
or model-based dominance.

The predecessor matrix supports checkpoint restoration, Ant fixed-route
avoidance, and a violation-cost tradeoff, but not separate universal effects of
confidence or predicates. Sparse residuals use fewer parameters and export
per-run equations, yet show no endpoint advantage, mixed calibration error, and
low cross-seed support overlap. Inspectability holds; stable structural
discovery and sparse control superiority do not.

\subsection{Limitations}

The evidence covers three controlled MuJoCo shifts, not physical sim-to-real
transfer or within-task route switching. DARC is the only external adaptation
method under the identical access contract. Every adaptive method receives
50,000 target training interactions plus validation and test access, and the
candidate may use validation for actor restoration; reduced total access and
one-shot adaptation are not established. Confidence and predicates are
empirical weights, not uncertainty bounds or safety guarantees. The Ant
observation omits absolute planar position, although its radial reward and wall
termination depend on that position. Its observation-space MDP is therefore an
approximation: identical model inputs can have different rewards or wall
outcomes, limiting the reward and termination heads. Broader shifts,
matched adaptation baselines, and physical evaluation remain necessary.

%% file: sections/statements.tex
\section*{Artificial Intelligence Use Statement}

Artificial intelligence (AI) assistance through OpenAI Codex was used
for literature and citation checks, feedback on method and experiment design,
implementation audits and code editing,
experiment orchestration, statistical aggregation, result interpretation, and
manuscript drafting. The authors selected the research questions and frozen
protocol, approved the changes, and reviewed the source metadata, mathematical
statements, code, and reported artifacts. No large language model was called
inside the reported training or evaluation runs, generated an experimental
outcome, or selected which completed test results to report. The authors take
responsibility for the final text, claims, code, and artifacts.

\section*{Ethics Statement}

The study uses simulated control environments and does not involve human
participants, personal data, animals, or physical-robot deployment. Confidence
and predicate weights are empirical model-learning mechanisms and are not a
formal safety guarantee.

\section*{Reproducibility Statement}

Section~\ref{sec:experiments} defines the environment shifts, comparators,
interaction accounting, seeds, and statistical procedure. Appendix
\ref{app:implementation} lists the main hyperparameters, and Appendix
\ref{app:reproducibility} identifies the frozen source and the single command
that regenerates the validated submission outputs. The anonymous supplemental
archive contains source code, environment definitions, per-seed endpoint
values, validation curves, target-access ledgers, symbolic supports, figure
source data, sanitized per-run manifests and metrics, and the regeneration
scripts. The manifests preserve experimental provenance but replace the
machine-local run directory and Git revision; checkpoints are excluded because
no reported result requires them. The public code repository is
\url{https://github.com/erzhu419/MC-WM}.

%% file: sections/appendix.tex
\section{Complete Notation Ledger}
\label{app:notation}

Table~\ref{tab:notation-full} fixes one meaning for every nonstandard symbol
used in the manuscript. Standard arithmetic operators retain their usual
meaning. Indices are fixed as follows: $t$ is interaction time, $i$ is a data sample, $i'$
is an auxiliary sample in a normalization, $k$ is a joint prediction-output
coordinate ($1\leq k\leq n_s+1$), $j$ is a policy critic ($1\leq j\leq3$),
$m$ is a zero-based state coordinate ($0\leq m<n_s$), $n$ is a zero-based action
coordinate ($0\leq n<n_a$), $u$ is a validation-checkpoint index, $\mathfrak m$ is a candidate
world-model family, and $\ell$ is a domain.
Generic $s$, $a$, and $x$ omit the interaction index; $s'$ denotes a successor
state. Hats denote predictions, and the superscript $(\mathrm{R})$ restricts
a prediction to the residual candidate. Parameters retain the same roles:
$\phi$ for the policy, $\psi$ for the neural state remainder, $\zeta$ for the
reward correction, $\nu$ for the termination head, $\xi$ for the confidence
critic, and $\eta_j$ for policy critic $j$; $\bar\xi$ is the confidence target's
parameter vector. $\mathbb{R}$ denotes the real numbers,
$\lVert\cdot\rVert_2$ the Euclidean norm, $\odot$ elementwise multiplication,
$|\mathcal{I}_{\mathrm{cal}}|$ set cardinality, $|\cdot|$ on scalars absolute
value, and $\mathbb{1}[\cdot]$ the Boolean indicator.

\begin{longtable}{p{0.27\linewidth}p{0.67\linewidth}}
\caption{Complete mathematical notation.}\label{tab:notation-full}\\
\toprule
Symbol & Unique meaning \\
\midrule
\endfirsthead
\toprule
Symbol & Unique meaning \\
\midrule
\endhead
$\mathcal{S}$; $\mathcal{A}$ & state space; action space \\
$\mathcal{M}_{\ell}$ & domain MDP, with $\ell\in\{\mathrm{S},\mathrm{T}\}$ denoting simulator or target \\
$P_{\ell}$; $\rho_{\ell}$ & transition kernel; expected one-step reward in domain $\ell$ \\
$\gamma$; $p_0$ & policy-return discount; initial-state distribution \\
$\pi_{\phi}$ & policy with parameter vector $\phi$ \\
$J_{\mathcal{M}}(\pi)$ & discounted return of policy $\pi$ in MDP $\mathcal{M}$ \\
$\mathbb{E}$ & expectation under the subscripted process \\
$s_t$; $a_t$; $r_t$; $\kappa_t$ & observed state; action; observed reward; observed termination indicator at time $t$ \\
$s^{(m)}$; $a^{(n)}$ & coordinate $m$ of a generic state; coordinate $n$ of a generic action \\
$x_t=[s_t;a_t]$ & concatenated state-action input \\
$n_s$; $n_a$; $p$ & state dimension; action dimension; deployed feature count \\
$\mathcal{D}_{\mathrm{S}}$; $\mathcal{D}_{\mathrm{T}}$; $\mathcal{D}_{\mathrm{T}}^{0}$ & simulator training dataset; target training dataset; initial target dataset available before online interaction \\
$\mathcal{I}_{\mathrm{fit}}$; $\mathcal{I}_{\mathrm{sel}}$; $\mathcal{I}_{\mathrm{cal}}$ & initial target fit; feature-selection and early-stopping; calibration index set \\
$\widehat{M}_{\mathrm{S}}$; $\widehat{M}_{\mathrm{R}}$; $\widehat{M}_{\mathrm{D}}$; $\widehat{M}_{\mathrm{T}}$ & frozen simulator model; residual candidate; direct target candidate; deployed target model \\
$z_i$; $\widehat z_i^{(\mathfrak m)}$ & joint state-increment and reward target; its prediction from candidate family $\mathfrak m$ \\
$\varsigma_k$ & fit-partition scale for joint-target coordinate $k$ \\
$\mathfrak m$; $\mathfrak m^{\star}$ & candidate world-model family; deployed family \\
$\mathcal{R}_{\mathfrak m}$ & standardized calibration transition risk for family $\mathfrak m$ \\
$\widehat u_{\mathrm{S}}(x)$; $\widehat\rho_{\mathrm{S}}(x)$ & simulator state-increment; reward prediction \\
$\widehat\Delta_s(x)$; $\widehat\Delta_{r,\zeta}(x)$ & target state; reward correction, with reward parameters $\zeta$ \\
$\widehat\Delta_{s,\mathrm{sp}}^{(m)}(s,a)$ & coordinate $m$ of the reported sparse state-correction branch \\
$\widehat p_{\nu}(s,a,s')$ & termination probability with parameters $\nu$ \\
$\widehat s_{t+1}$; $\widehat r_t$; $\widehat\kappa_t$ & deployed-model state; reward; termination prediction \\
$y_i^{\Delta s}$ & target state-correction regression target for sample $i$ \\
$\mathcal{F}$; $\Theta(x)$; $\widetilde\Theta(x)$ & candidate feature set with fixed evaluation order; its unscaled evaluation vector; scaled deployed feature vector \\
$B$; $\mu$ & sparse state-correction coefficient matrix; binary feature-support mask \\
$g_{\psi}$ & masked neural state remainder with parameters $\psi$ \\
$f$; $\mathcal{L}_{\mathrm{sel}}$; $\epsilon_{\mathrm{feat}}$ & candidate feature; selection loss; required selection improvement \\
$\omega_i$; $d_i$; $h_d$ & policy-visitation refit weight; root mean squared standardized Euclidean distance to eight neighbors; positive distance bandwidth \\
$e_i$; $\tau_e$; $c_i$ & state prediction MSE; running median-error scale; one-step confidence target \\
$\boldsymbol{\sigma}_{\mathrm{ens}}(s,a)$ & coordinatewise state-increment and reward standard deviation from the route-specific uncertainty ensemble \\
$Q^{\Delta}_{\xi}$; $Q^{\Delta}_{\bar\xi}$ & log-confidence critic; Polyak-averaged target critic \\
$a_{i+1}^{\mathrm{det}}$ & deterministic action of the current policy at sample $i$'s successor \\
$\gamma_{\Delta}$; $q_i^{\mathrm{tgt}}$ & confidence discount; confidence regression target \\
$w_i^{\Delta}$ & imagined-transition confidence weight \\
$\mathcal{C}$; $\chi_i$ & deterministic predicate set; binary predicate-pass indicator \\
$w_i$ & final policy-critic loss weight for transition $i$ \\
$\mathcal{I}$; $b$ & policy-optimization minibatch index set; its size \\
$Q_{\eta_j}$; $y_i^Q$ & policy critic $j$ with parameters $\eta_j$; standard SAC target \\
$\mathcal{L}_Q$; $\lambda_{\mathrm{ens}}$ & weighted policy-critic loss; critic-disagreement coefficient \\
$\iota_i$; $\lambda_{\mathrm{v}}(t)$; $r_i^{\mathrm{train}}$ & binary violation indicator; online violation cost; reward consumed by policy training \\
$\varrho_{\mathrm{cal}}$ & initial simulator-to-residual state RMSE ratio on the calibration partition \\
$\overline G_u$; $\overline V_u$ & mean undiscounted validation return; mean validation violations per episode at checkpoint $u$ \\
$\overline G_{\mathrm{peak}}$; $\overline G_{\mathrm{final}}$ & peak; final-checkpoint mean undiscounted validation return \\
$K$; $N_{\mathrm{on}}$ & member count in either probabilistic world-model ensemble; online target-training budget \\
$q_z$; $v_x$ & HalfCheetah root vertical generalized coordinate; Ant forward velocity \\
\bottomrule
\end{longtable}

\section{Auditability Diagnostics}
\label{app:auditability}

Table~\ref{tab:auditability} reports diagnostics from the ten complete MC-WM
runs in each environment. Support and state-error columns describe the residual
candidate even when the direct family is deployed: the initial fit on Ant and
gravity, and the final residual refit on Walker. SINDy MSE and additive MSE are
unstandardized state-correction errors on that fit's calibration partition;
the error 95th percentile is the Euclidean norm of the additive state error,
not the standardized joint routing risk in Equation~\eqref{eq:routing-risk}.
The support Jaccard statistic compares active
coordinate-feature pairs between every pair of training seeds. Low overlap is
evidence against cross-seed structural stability; it does not prevent a run's
locked equation and coefficients from being inspected. Confidence is evaluated
on the fixed target-replay prefix defined in Section~\ref{sec:experiments}; the
rejection rate aggregates every predicate check made during imagined rollouts.

\begin{table}[ht]
\caption{Per-run model diagnostics aggregated across training seeds. Parentheses
contain cross-seed SD unless a range is shown.}
\label{tab:auditability}
\centering
\scriptsize
\resizebox{\linewidth}{!}{\input{generated/final_confirmation/auditability/auditability_diagnostics.tex}}
\end{table}

Equations~\eqref{eq:sparse-ant_wall_broken_soft_ceiling-example},
\eqref{eq:sparse-gravity_soft_ceiling-example}, and
\eqref{eq:sparse-friction_walker_soft_ceiling-example} give
the complete, untruncated representative coordinate corrections selected by the
fixed rule below. They report only $B\widetilde\Theta$, not the neural remainder
in Equation~\eqref{eq:additive-correction}:
\input{generated/final_confirmation/auditability/auditability_equation_examples.tex}
The supplemental auditability directory contains \path{full_symbolic_support.json}, which
stores every active term and coefficient for all thirty final-candidate runs.

For display only, the representative seed maximizes mean support Jaccard to the
other nine seeds. Within that run, the fixed rule prefers nonempty coordinates
with at most eight terms, then orders them by cross-seed coordinate support
Jaccard, fewer terms, and output label. This rule does not affect model fitting,
route selection, or endpoint analysis.

\section{Registered Mechanism and Residual Studies}
\label{app:ablations}

\subsection{Exact routing counterfactual}
\label{app:exact-routing}

On the fresh confirmation seeds, all three arms execute the same residual and
direct-target candidate fitting code with matched data, initialization, and
isolated component random streams. Direct-candidate fitting saves and restores
the global NumPy and PyTorch random states, including visible accelerators;
its dedicated stream is derived deterministically from the training seed.
The automatic arm deploys the lower-risk
family; the controls force residual or direct deployment only after both risks
have been recorded. Violation costs, trust filters, actor updates, termination
treatment, and checkpoint restoration are identical. The analyzer requires
equal non-routing arguments and diagnostics frozen before deployment; mutable
online residual-refit diagnostics are excluded. For each seed, the automatic
arm must also have a byte-identical metric stream and equal summary, after
removing two route-label fields, to the forced arm for its selected family.

\subsection{Predecessor mechanism matrix}
\label{app:predecessor}

A source-identical $3\times8\times10$ matrix starts from the frozen gap-routed
predecessor and removes model-family routing, confidence weighting,
deterministic predicates, the violation cost, validation checkpoint
restoration, or the sparse state branch. A seventh control removes both routing
and the sparse branch, allowing a fixed-residual sparse contrast. This matrix
tests the predecessor; it is not relabelled as an ablation of the final
calibration-risk selector.

\begin{table}[ht]
\caption{Paired component matrix for the frozen gap-routed predecessor.
Differences are ablation minus predecessor; negative return therefore favors the
predecessor, while negative violations favor the ablation. This matrix is not a
source-identical ablation of the final calibration-risk selector.}
\label{tab:mechanism-ablations}
\centering
\scriptsize
\resizebox{\linewidth}{!}{\input{generated/mechanism_confirmation_results.tex}}
\end{table}

\subsection{Sparse versus dense state residual}
\label{app:sparse-dense}

A fresh $3\times2\times10$ matrix compares the sparse additive residual plus
masked neural remainder with a two-hidden-layer dense neural residual. Both
arms force residual deployment and match target access, source-model data,
reward and termination heads, trust filters, policy updates, and evaluation.
The dense arm uses standardized raw state-action inputs and performs no feature
discovery or symbolic regression. Primary endpoints are return and violations;
calibration transition error and state-residual parameter count are secondary.

\begin{table}[ht]
\caption{Paired dense-minus-sparse residual comparison under fixed residual
deployment. Calibration risk is standardized transition error, and state
parameters count trainable parameters in the state-correction branch.}
\label{tab:sparse-dense}
\centering
\scriptsize
\resizebox{\linewidth}{!}{\input{generated/sparse_dense_residual_results.tex}}
\end{table}

\section{Target-Access Ledger and Learning Curves}
\label{app:target-access}

\begin{figure}[ht]
\centering
\includegraphics[width=\linewidth]{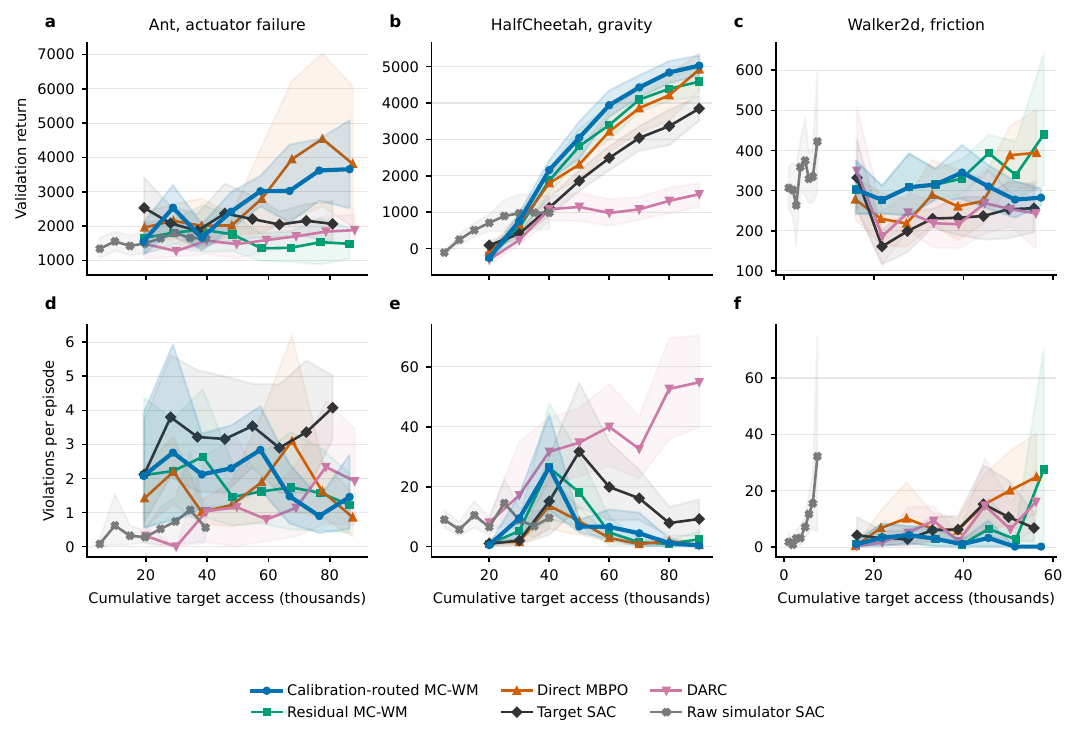}
\caption{Validation return and violations against cumulative target access.
Lines are ten-seed means and bands are 20,000-resample percentile-bootstrap
95\% confidence intervals for the cross-seed mean. The horizontal coordinate
includes target training plus cumulative validation interactions. Validation
outcomes do not enter gradient updates, but the registered candidate may use
its validation trajectory for checkpoint restoration.}
\label{fig:learning-curves}
\end{figure}

\begin{figure}[ht]
\centering
\includegraphics[width=\linewidth]{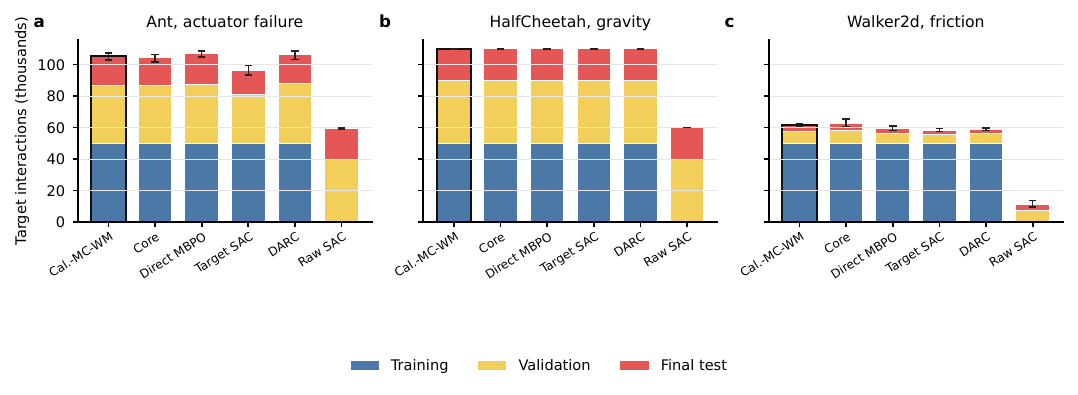}
\caption{Actual target access by method and environment. Bars show cross-seed
means for training, validation, and final-test transitions; error bars show the
20,000-resample percentile-bootstrap 95\% CI of mean total access.}
\label{fig:target-ledger}
\end{figure}

Figure~\ref{fig:target-ledger} separates target training, validation, and final
test interactions. Figure~\ref{fig:learning-curves} plots
validation return against cumulative target access through that checkpoint;
its horizontal axis therefore includes validation interactions. The registered
candidate does not use validation outcomes for gradient updates, but it can use
them to select the actor evaluated at the final test.

\section{Implementation Details}
\label{app:implementation}

Table~\ref{tab:hyperparameters} lists the fixed calibration-routed candidate
settings.

\begin{table}[ht]
\caption{Fixed MC-WM hyperparameters used by every calibration-routed candidate run.}
\label{tab:hyperparameters}
\centering
\small
\resizebox{\linewidth}{!}{%
\begin{tabular}{lll}
\toprule
Component & Parameter & Value \\
\midrule
Simulator ensemble & members / hidden layers / width & $5/3/200$ \\
Simulator ensemble & batch size / learning rate & $256/10^{-3}$ \\
Residual split & fit / selection / calibration & $0.8/0.1/0.1$ \\
Model-family route & metric / tie rule & standardized joint RMSE / residual \\
Sparse regression & ridge / threshold & $0.05/0.05$ \\
Feature discovery & budget / minimum gain & $100/10^{-5}$ \\
Neural remainder & hidden width / learning rate & $32/10^{-3}$ \\
Reward and termination heads & hidden width & 64 \\
Online deployed-model refit & frequency / maximum rows & $1{,}000/50{,}000$ \\
Imagined data & horizon / batch / frequency & $1/400/250$ \\
Target / imagined replay capacity & transitions & $100{,}000/50{,}000$ \\
Confidence critic & hidden width / $\gamma_{\Delta}$ & $128/0.5$ \\
Policy learner & critics / hidden width & $3/256$ \\
Policy learner & $\gamma$ / learning rate & $0.99/(3\times10^{-4})$ \\
Policy learner & pessimism / disagreement coefficient & $-2.0/0.01$ \\
Online exploration & uniform-action cutoff & $t<2{,}000$ \\
Actor route & mixed-update calibration band & $[2.0,4.5)$ \\
Mixed actor & synthetic trust power / update interval & $0.5/4$ critic updates \\
Standard actor & update interval & 2 critic updates \\
Violation cost & start / final value & $0.5N_{\mathrm{on}}/10$ \\
Validation actor selection & violation budget / return retention & $1.0/0.8$ \\
\bottomrule
\end{tabular}
}
\end{table}

\subsection{Residual fitting details}
\label{app:residual-details}

STLSQ alternates ridge regression and coefficient thresholding, both set to
0.05. An additional deterministic orthogonal feature $f$ is accepted only when
\begin{equation}
\mathcal{L}_{\mathrm{sel}}(\mathcal{F}\cup\{f\})
<\mathcal{L}_{\mathrm{sel}}(\mathcal{F})-\epsilon_{\mathrm{feat}},
\qquad \epsilon_{\mathrm{feat}}=10^{-5},
\label{eq:feature-gate}
\end{equation}
where an admitted $f$ is appended to the fixed feature order, and
$\mathcal{L}_{\mathrm{sel}}$ is the selection-partition MSE after
refitting on $\mathcal{I}_{\mathrm{fit}}$. Before this gate, an orthogonal
candidate is discarded when its largest estimated absolute coefficient exceeds
5. This is a local stability screen, not a bound on the complete coefficient
matrix. The feature budget is $p\leq100$; Equation~\eqref{eq:feature-gate} is
the only selection-loss acceptance rule.

After discovery, the feature order and support mask $\mu$ are locked.
Residual-route refits update $B$, $g_{\psi}$, and the reward and termination
heads without adding features. The latest 10,000 state-action pairs form the
policy reference bank; the distance calculation uses a reproducible 8,000-row
subsample when the bank is larger. For refit sample $i$, let $d_i$ be the
square root of the mean squared standardized Euclidean distance to its eight
nearest retained reference points (with $10^{-8}$ added inside the square
root), and let $h_d$ be the median of these distances plus $10^{-6}$.
Its refit weight is
\begin{equation}
\omega_i=0.1+0.9\,
\frac{\exp(-d_i/h_d)}{\max_{i'}\exp(-d_{i'}/h_d)}.
\label{eq:policy-weight}
\end{equation}
The index $i'$ ranges over the refit samples. Equation~\eqref{eq:policy-weight}
weights the sparse and neural state losses; it is not an importance ratio or a
value-error bound.

A separate 64-unit ReLU classifier estimates target termination probability
$\widehat p_{\nu}(s,a,s')$. The deployed termination prediction is
\begin{equation}
\widehat\kappa_t=
\mathbb{1}\!\left[
\widehat p_{\nu}(s_t,a_t,\widehat s_{t+1})\geq0.5
\right],
\label{eq:corrected-termination}
\end{equation}
where $\nu$ denotes classifier parameters and $\mathbb{1}[\cdot]$ is the
indicator function. The initially fitted head is carried onto either route;
residual-route refits update it. Equation~\eqref{eq:corrected-termination} is
consumed only when
the observed target-replay termination rate exceeds 2\%, avoiding an
effectively unidentified rare-event classifier.

\subsection{Actor routing, checkpoint selection, and complete procedure}
\label{app:policy-details}

The initial state-only calibration ratio $\varrho_{\mathrm{cal}}$ divides the
frozen simulator state RMSE by the residual candidate state RMSE on
$\mathcal{I}_{\mathrm{cal}}$. When
$2\leq\varrho_{\mathrm{cal}}<4.5$, actor updates use equal numbers of target and
imagined states; imagined actor samples carry weight $(w_i)^{1/2}$, and the
actor updates after every four critic updates. Outside this band, target and
imagined replay use the standard separate SAC updates.

At each 5,000-step checkpoint, five fixed validation episodes yield
mean undiscounted return $\overline G_u$ and mean violations per episode
$\overline V_u$. MC-WM tracks
the highest-return actor satisfying $\overline V_u\leq1$. Before the one-time
test, that actor replaces the final actor only if either (i) final validation
return is below the peak by more than
$0.2\max(|\overline G_{\mathrm{peak}}|,1)$ and the tracked actor has higher
return, or (ii) final validation violations exceed one and the tracked return
is at least the final return minus
$0.2\max(|\overline G_{\mathrm{final}}|,1)$. Here
$\overline G_{\mathrm{peak}}=\max_u\overline G_u$ is the maximum over all
validation checkpoints, including ineligible ones, and
$\overline G_{\mathrm{final}}$ is the final-checkpoint mean. The critics and world model are
not selected. Final testing uses 20 fixed seeds disjoint from training and
validation, exactly once after any actor restoration.

\begin{algorithm}[ht]
\caption{Calibration-routed MC-WM training and held-out testing}
\label{alg:mcwm}
\begin{algorithmic}[1]
\Require simulator data $\mathcal{D}_{\mathrm{S}}$, initial target data $\mathcal{D}_{\mathrm{T}}^{0}$, online target budget $N_{\mathrm{on}}$
\State Fit and freeze the simulator ensemble $\widehat M_{\mathrm{S}}$
\State Split $\mathcal{D}_{\mathrm{T}}^{0}$ into $\mathcal{I}_{\mathrm{fit}}$, $\mathcal{I}_{\mathrm{sel}}$, and $\mathcal{I}_{\mathrm{cal}}$
\State Fit $\widehat M_{\mathrm{R}}$ and fit $\widehat M_{\mathrm{D}}$ in an isolated component random stream
\State Compute $\mathcal{R}_{\mathrm{R}}$, $\mathcal{R}_{\mathrm{D}}$, and $\varrho_{\mathrm{cal}}$; fix $\mathfrak m^{\star}$ and the actor-update route
\State Lock residual feature order and support $\mu$; initialize target replay, imagined replay, $\pi_{\phi}$, and $Q^{\Delta}_{\xi}$
\State Fit $Q^{\Delta}_{\xi}$ on initial target-replay prediction errors
\For{$t=1,\ldots,N_{\mathrm{on}}$}
  \State Execute one target action and append the observed transition
  \If{$t$ is a 1,000-step model-refit point}
    \State Warm-start refit only the deployed state-and-reward model
    \State Update $Q^{\Delta}_{\xi}$ from target-replay prediction errors
  \EndIf
  \If{$t$ is a 250-step imagination point}
    \State Generate one-step transitions from $\widehat M_{\mathrm{T}}$
    \State Compute $w_i^{\Delta}$, $\chi_i$, $w_i$, and $\iota_i$; retain rows with $w_i>0.01$
  \EndIf
  \State Apply $\lambda_{\mathrm{v}}(t)$ and update SAC using the fixed actor route
  \If{$t$ is a 5,000-step validation point}
    \State Record $\overline G_u$ and $\overline V_u$; update the best eligible actor checkpoint
  \EndIf
\EndFor
\State Apply the registered trajectory restoration rule, then evaluate once on held-out test seeds
\end{algorithmic}
\end{algorithm}

\paragraph{Direct target MBPO.}
This comparator fits the same five-member, three-layer, width-200 probabilistic
ensemble directly to target transitions. It uses the same policy learner,
one-step rollout size and frequency, and target and imagined replay capacities
as MC-WM, but has no simulator pretraining, residual, confidence weight, or
predicate. The ensemble fits for at most 100 epochs initially and is refit
every 1,000 online steps for at most 20 epochs on at most 50,000 target
transitions; both stages use a 10\% internal holdout and early stopping.

\paragraph{Dense neural residual control.}
The frozen simulator receives a dense state-correction network with two
width-64 ReLU hidden layers. Raw state-action inputs and state-correction targets
are standardized from the initial fit partition; the disjoint selection
partition supplies early stopping, and the calibration partition remains
untouched. Adam uses learning rate $10^{-3}$, weight decay $10^{-4}$, and batch
size 256. The reward and termination heads, policy-aware refit weights, trust
filters, policy learner, and target-access schedule are identical to the sparse
fixed-route arm. The dense arm performs no feature discovery and exports no
symbolic coefficients.

\paragraph{Matched DARC implementation.}
The DARC comparator uses two binary domain classifiers. The first receives
$(s_t,a_t)$; the second has an additive state-action logit and a conditional
next-state logit for $(s_t,a_t,s_{t+1})$. Each hidden layer has width 32 and
classifier inputs receive Gaussian noise with SD 1.0. Source rewards are
adjusted by the difference between target-versus-source transition and
state-action logits. Classifiers train from the same 50,000 source and 50,000
target training interactions available under the matched protocol. The
4,000-step correction warmup is disclosed because it differs from the original
long-horizon schedule.

\section{Exact Environment Transformations}
\label{app:environments}

All modifications below are applied identically to training, validation, and
test instances of the corresponding domain. In the HalfCheetah gravity
environment, the simulator's vertical gravity is set to
$-19.62\,\mathrm{m\,s^{-2}}$ and the target retains the MuJoCo default
$-9.81\,\mathrm{m\,s^{-2}}$. Both members continue the episode when $q_z>0.2$
and subtract
$10[\max(0,q_z-0.2)]^2$ from the standard HalfCheetah reward.

For Friction Walker, every sliding, torsional, and rolling geom-friction
coefficient is multiplied by 0.3 in the simulator; the target retains the
default coefficients. Both domains subtract
$10[\max(0,s^{(0)}-1.25)]^2$ from the standard Walker2d reward, where
$s^{(0)}$ is the root-height observation.

For Ant actuator failure, target actions 4--7 are set to zero before the
physics step; the simulator applies all eight actions. Both domains retain the
standard Ant health termination, add a rear-wall termination when the
world-frame base position falls below $-3$ along the first horizontal coordinate,
and set the wall-step reward to zero. Away from the wall, the project-specific
reward is planar Euclidean distance from the origin plus 1, minus 0.5 times the
squared norm of the applied action. The soft cap further subtracts
$[\max(0,|v_x|-1.5)]^2$. Thus disabled target channels incur no control cost
after they are zeroed. The reported Ant method has an empty deterministic
predicate set; these environment rewards and terminations still apply.

\section{Reproducibility and Frozen Source}
\label{app:reproducibility}

The final main matrix and exact forced-route controls record source-tree
identity:
\par\noindent
\nolinkurl{939a71c718b4f8d2296a41c5c43cd8323e84fdedfd7ce374104d05ae420af9c0}.
\par
The predecessor mechanism matrix records:
\par\noindent
\nolinkurl{66eb8a36ee482d3a22000b8302c91c8da4ac5914f59bbd14e9182f581f028a8e}.
\par
The sparse-versus-dense study records:
\par\noindent
\nolinkurl{06e1d3eea255e83c7be769e364ebc59a2e63a4ee1feb473a81ecec24bf52853c}.
\par
Git revisions are withheld during double-blind review; later manuscript and
packaging changes leave these experimental methods unchanged. One Ant
forced-residual routing cell (seed 976484671) replaces task t88897 with t89031
after three pre-deployment diagnostics failed equality across nodes. The
replacement used identical source, arguments, method, and seed on the
candidate's node and passed the exact gate. It was registered before completion
or inspection of aggregated endpoints. The ledger contains 540 unique cells
from 541 executions: 180 main comparisons, 60 additional route controls, 240
predecessor cells, and 60 sparse-versus-dense cells; reused candidates count
once. The archive includes anonymized frozen sources, runnable code, sanitized
per-run JavaScript Object Notation (JSON) and JSON Lines (JSONL) artifacts,
figure source data, and per-run arguments, dependencies, seeds, metrics, and
interaction ledgers. Run \path{experiments/build_submission_outputs.sh} to
regenerate assets from complete protocol-matched seed sets. No external
language-model service is used by any reported matrix.

%% file: generated/final_confirmation/auditability/auditability_diagnostics.tex
\begin{tabular}{lrrrrrrr}
\toprule
Environment & Active coeff. & Support overlap & SINDy MSE & Additive MSE & Error 95th pct. & Mean confidence & Reject rate \\
\midrule
Ant, actuator failure & 228.9 (50.5) & 0.24 [0.19, 0.30] & 0.34 (0.04) & 0.21 (0.06) & 4.32 (0.73) & 0.23 (0.08) & 0.0\% (0.0) \\
HalfCheetah, soft gravity & 271.2 (58.9) & 0.26 [0.16, 0.37] & 0.98 (0.94) & 0.40 (0.05) & 5.24 (0.32) & 0.27 (0.02) & 35.1\% (8.4) \\
Walker2d, friction & 210.5 (30.2) & 0.20 [0.15, 0.26] & 2.53 (1.07) & 0.56 (0.20) & 6.22 (1.01) & 0.26 (0.04) & 55.9\% (4.5) \\
\bottomrule
\end{tabular}

%% file: generated/final_confirmation/auditability/auditability_equation_examples.tex
\paragraph{Ant, actuator failure (support-medoid seed 952414860).}
\begin{equation}
\begin{aligned}
\widehat{\Delta}_{s,\mathrm{sp}}^{(9)}(s,a) &= -0.138\,a^{(6)}
\end{aligned}
\label{eq:sparse-ant_wall_broken_soft_ceiling-example}
\end{equation}
\paragraph{HalfCheetah, soft gravity (support-medoid seed 952414860).}
\begin{equation}
\begin{aligned}
\widehat{\Delta}_{s,\mathrm{sp}}^{(7)}(s,a) &= -0.128\,s^{(0)}s^{(1)}
\end{aligned}
\label{eq:sparse-gravity_soft_ceiling-example}
\end{equation}
\paragraph{Walker2d, friction (support-medoid seed 295539546).}
\begin{equation}
\begin{aligned}
\widehat{\Delta}_{s,\mathrm{sp}}^{(9)}(s,a) &= -8.6 + 7.04\,s^{(0)} - 9.02\,s^{(1)} + 2.04\,s^{(2)} \\
&\quad + 0.943\,s^{(3)} - 1.07\,s^{(5)} + 4.2\,s^{(9)} - 0.0474\,s^{(15)} \\
&\quad - 0.0382\,s^{(16)} + 3.74\,s^{(0)}s^{(1)} - 0.143\,s^{(0)}s^{(2)} + 0.399\,s^{(0)}s^{(5)} \\
&\quad - 2.22\,s^{(0)}s^{(9)} + 0.033\,s^{(0)}s^{(15)} + 0.0574\,s^{(0)}s^{(16)} - 0.44\,s^{(1)}s^{(3)}s^{(6)} \\
&\quad + 1.65\,s^{(0)}s^{(3)}s^{(6)}
\end{aligned}
\label{eq:sparse-friction_walker_soft_ceiling-example}
\end{equation}

%% file: generated/mechanism_confirmation_results.tex
\begin{tabular}{llrr}
\toprule
Environment & Ablation & Return difference (95\% CI) & Violation difference (95\% CI) \\
\midrule
Ant actuator & No checkpoint selection & -428.8 [-650.1, -212.8] & -0.01 [-0.32, 0.33] \\
Ant actuator & No confidence weighting & 593.8 [-2024.7, 3881.6] & 2.28 [-0.38, 6.29] \\
Ant actuator & Fixed residual route & -2231.8 [-4461.5, -871.3] & -0.50 [-1.18, 0.17] \\
Ant actuator & Fixed residual route, neural only & -2302.4 [-4409.3, -901.8] & -0.57 [-1.23, 0.02] \\
Ant actuator & No predicate filtering & -162.2 [-2178.5, 1960.5] & 1.43 [-0.72, 4.96] \\
Ant actuator & No sparse branch & 0.0 [0.0, 0.0] & 0.00 [0.00, 0.00] \\
Ant actuator & No violation cost & 1560.0 [-960.7, 4793.5] & 17.02 [0.83, 46.07] \\
HalfCheetah gravity & No checkpoint selection & 57.1 [0.0, 171.3] & 0.17 [0.00, 0.52] \\
HalfCheetah gravity & No confidence weighting & -150.6 [-809.6, 685.1] & -0.15 [-0.93, 0.52] \\
HalfCheetah gravity & Fixed residual route & -202.1 [-477.8, 0.0] & -0.25 [-0.62, 0.00] \\
HalfCheetah gravity & Fixed residual route, neural only & -232.2 [-703.1, 287.4] & 0.08 [-0.77, 0.77] \\
HalfCheetah gravity & No predicate filtering & -332.9 [-888.4, 230.9] & 0.17 [-0.79, 1.09] \\
HalfCheetah gravity & No sparse branch & -175.0 [-599.0, 316.5] & -0.06 [-0.88, 0.64] \\
HalfCheetah gravity & No violation cost & 89.2 [-197.2, 411.0] & 2.55 [-0.17, 6.33] \\
Walker friction & No checkpoint selection & -84.2 [-158.8, -24.9] & -0.78 [-2.32, 0.01] \\
Walker friction & No confidence weighting & -11.3 [-87.1, 71.5] & -0.76 [-2.30, 0.03] \\
Walker friction & Fixed residual route & 0.0 [0.0, 0.0] & 0.00 [0.00, 0.00] \\
Walker friction & Fixed residual route, neural only & -26.6 [-188.1, 150.9] & -0.68 [-2.24, 0.13] \\
Walker friction & No predicate filtering & -6.2 [-173.7, 167.1] & -0.69 [-2.15, 0.07] \\
Walker friction & No sparse branch & -37.9 [-193.7, 135.6] & -0.69 [-2.24, 0.12] \\
Walker friction & No violation cost & 58.6 [6.7, 118.9] & 0.12 [0.03, 0.24] \\
\bottomrule
\end{tabular}

%% file: generated/sparse_dense_residual_results.tex
\begin{tabular}{lrrrrl}
\toprule
Environment & Return difference (95\% CI) & Violation difference (95\% CI) & Calibration-risk difference (95\% CI) & State-parameter difference (95\% CI) & Classification \\
\midrule
Ant actuator & 29.7 [-546.1, 652.8] & -0.04 [-0.81, 0.72] & 0.629 [0.485, 0.796] & 1414 [1396, 1431] & Unresolved \\
HalfCheetah gravity & 156.3 [-192.8, 506.9] & 0.66 [-0.24, 2.18] & -0.103 [-0.170, -0.047] & 1347 [1308, 1401] & Unresolved \\
Walker friction & 45.4 [-56.3, 147.4] & 0.33 [-0.08, 0.99] & -0.007 [-0.010, -0.004] & 1386 [1337, 1440] & Unresolved \\
\bottomrule
\end{tabular}